\documentclass[11pt]{article}

\usepackage[final]{acl}

\usepackage{times}
\usepackage{latexsym}

\usepackage[T1]{fontenc}

\usepackage[utf8]{inputenc}

\usepackage{microtype}

\usepackage{inconsolata}

\usepackage{graphicx}

\usepackage{algorithm}
\usepackage{algorithmic}

\usepackage{booktabs}

\usepackage{amsmath}
\usepackage{amssymb}

\usepackage{enumitem}

\usepackage{xcolor}

\usepackage{subcaption}

\usepackage{hyperref}

\usepackage{eso-pic}

\title{Sci-Reasoning: A Dataset Decoding AI Innovation Patterns}

\author{Jiachen Liu$^{*}$ \qquad Maestro Harmon$^{*}$ \qquad Zechen Zhang \\
  Orchestra Research \\
  \\
  $^{*}$Equal contribution
}

\begin{document}

\AddToShipoutPictureBG*{%
  \AtPageUpperLeft{%
    \put(30,-95){\includegraphics[width=2.5cm]{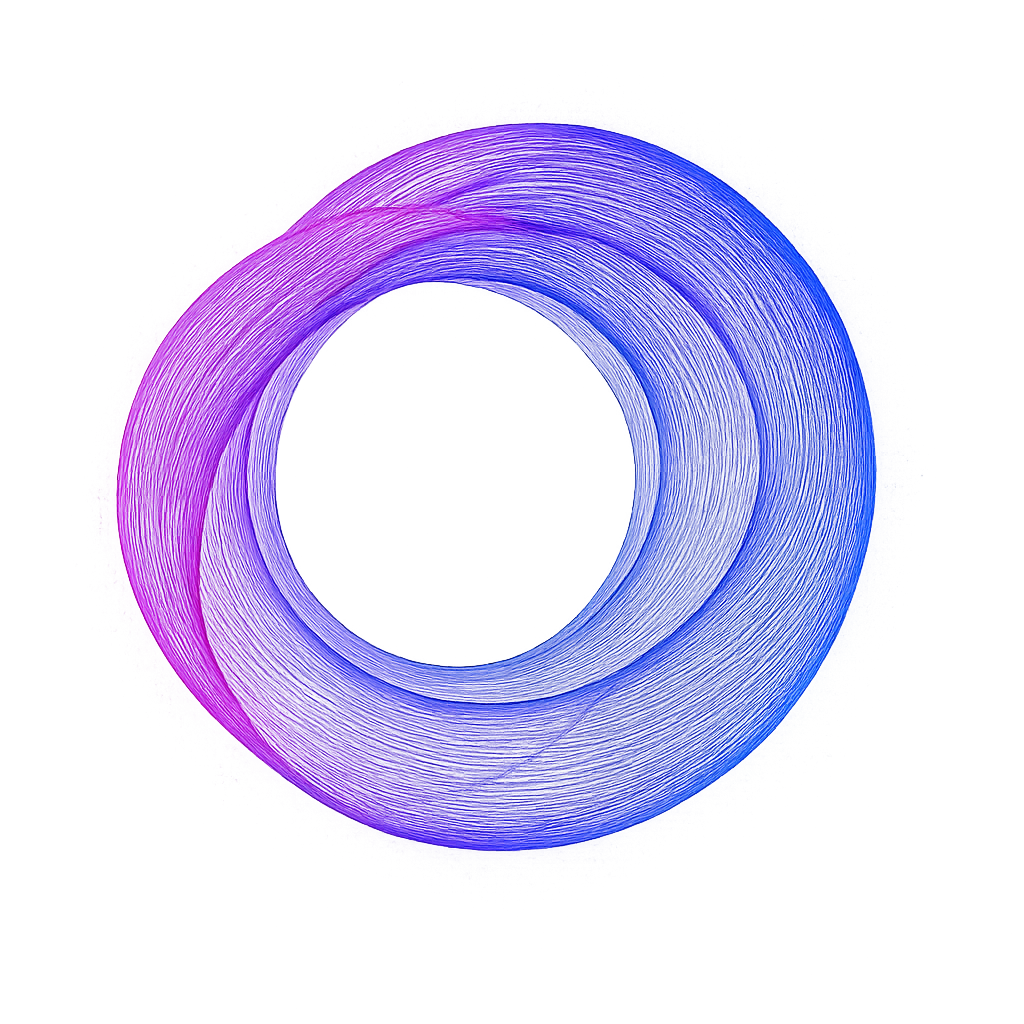}}%
  }%
}

\maketitle

\begin{abstract}
While AI innovation accelerates rapidly, the intellectual process behind breakthroughs---how researchers identify gaps, synthesize prior work, and generate insights---remains poorly understood. The lack of structured data on scientific reasoning hinders systematic analysis and development of AI research agents. We introduce \textbf{Sci-Reasoning}, the first dataset capturing the intellectual synthesis behind high-quality AI research. Using community-validated quality signals and an LLM-accelerated, human-verified pipeline, we trace Oral and Spotlight papers across NeurIPS, ICML, and ICLR (2023-2025) to its key predecessors, articulating specific reasoning links in a structured format. Our analysis identifies 15 distinct thinking patterns, with three dominant strategies accounting for 52.7\%: Gap-Driven Reframing (24.2\%), Cross-Domain Synthesis (18.0\%), and Representation Shift (10.5\%). The most powerful innovation recipes combine multiple patterns: Gap-Driven Reframing + Representation Shift,  Cross-Domain Synthesis + Representation Shift, and Gap-Driven Reframing + Cross-Domain Synthesis. This dataset enables quantitative studies of scientific progress and provides structured reasoning trajectories for training the next generation AI research agents.  
\end{abstract}

\section{Introduction}

The field of artificial intelligence is experiencing an unprecedented pace of innovation. Breakthroughs in large language models (LLMs), reinforcement learning from visual reasoning, vision-language-action models, and related AI systems have transformed what AI systems can accomplish~\cite{ouyang2022training,ramesh2022hierarchical,vaswani2017attention}. Yet despite this rapid progress, the intellectual process by which these breakthroughs emerge remains poorly understood. How do researchers identify promising gaps in existing work? How do they synthesize ideas from multiple predecessors into novel contributions? What patterns of reasoning characterize high-quality research? Understanding these reasoning trajectories is not only scientifically valuable in itself---it is the key to enabling the next generation of scientific discoveries. Yet currently, these questions are answered through subjective, anecdotal narratives rather than the structured data necessary for systematic analysis or machine learning.

\begin{figure*}[t]
\centering
\includegraphics[width=\textwidth]{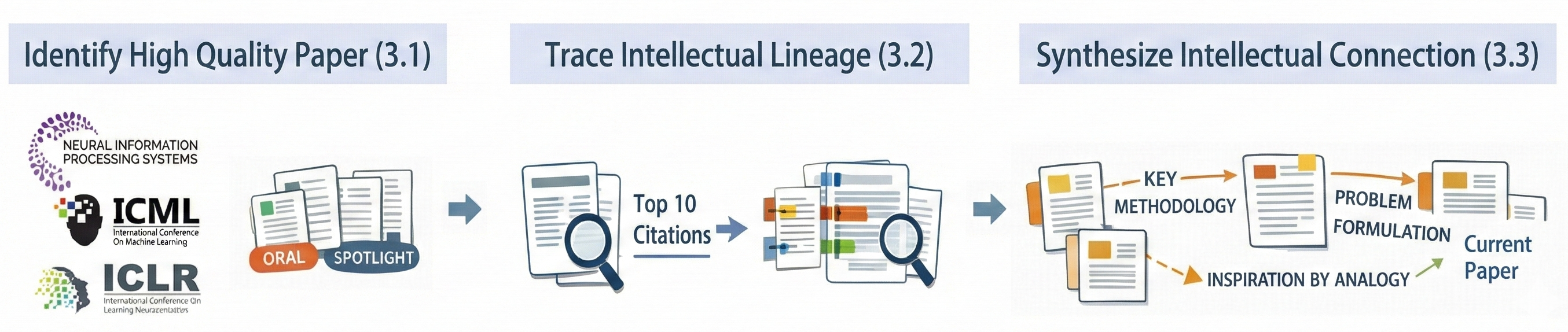}
\caption{Overview of the Sci-Reasoning dataset construction pipeline. Our methodology consists of four main stages: (1) identifying high-quality papers using community-validated signals, (2) tracing intellectual lineage to key predecessors via LLM analysis, (3) generating structured reasoning trajectories with lineage links that capture roles, relationships, and intellectual moves}
\label{fig:workflow}
\end{figure*}

The lack of structured, large-scale data on scientific \emph{reasoning}---the ``how'' and ``why'' behind building on prior work---prevents rigorous study of the innovation process itself. While citation networks provide valuable information about influence patterns~\cite{jo2022see,ghosal2021towards}, they capture only the \emph{fact} of a citation, not the \emph{nature} of the intellectual relationship. A paper might cite another work as a baseline to surpass, as a foundational concept to extend, or as a methodology to combine with other techniques. These distinctions are critical for understanding how breakthroughs happen, yet they are lost in simple citation graphs. Moreover, existing approaches to intellectual lineage tracing typically identify only a single ``progenitor'' paper~\cite{jo2022see} or treat citations uniformly~\cite{zhang2024pst}, missing the reality that most advances emerge from synthesizing insights across multiple prior works.

We introduce \textbf{Sci-Reasoning}, the first dataset designed to capture the structured intellectual synthesis behind high-quality AI research. Our contribution is three-fold: \textbf{(1)} We present a systematic methodology for identifying high-quality papers using community-validated signals (Oral/Spotlight status at NeurIPS, ICML, and ICLR) and tracing their intellectual lineage to its key predecessors through an LLM-accelerated, human-verified pipeline. Through a model ablation study comparing GPT-5.2, GPT-5, GPT-5-mini, and GPT-4.1 on predecessor extraction (Section~\ref{sec:model-ablation}), we identify GPT-5 as achieving the optimal cost-quality balance with 89.73\% recall, outperforming even newer models while maintaining computational efficiency for large-scale dataset construction. \textbf{(2)} We curate a dataset of 3,819 papers (999 Oral, 2,820 Spotlight) across NeurIPS, ICML, and ICLR (2023-2025) with richly annotated reasoning trajectories that capture not just \emph{which} papers influenced each breakthrough, but \emph{how}---through structured ``Lineage Links'' that include predecessor roles, relationship types, and natural language descriptions of the reasoning process. This structured format provides training data for AI research agents to learn expert research reasoning patterns. \textbf{(3)} We provide rigorous quality validation through test set evaluation against ground-truth papers and multi-model cross-validation (Section~\ref{sec:quality-validation}), demonstrating that frontier LLMs can predict research directions from intellectual predecessors with up to 49.35\% accuracy (Section~\ref{sec:experiments}). Figure~\ref{fig:workflow} provides an overview of our complete methodology pipeline.

Analysis of Sci-Reasoning reveals concrete, actionable patterns of innovation in AI research---patterns that represent the key to understanding and potentially automating aspects of future scientific discoveries. We identify \textbf{15 distinct thinking patterns}, with three dominant strategies accounting for 52.7\% of all papers. \textbf{Gap-Driven Reframing is the dominant thinking pattern} (924 papers, 24.2\%), where researchers diagnose a specific limitation and reframe the problem to map onto better-suited methods---suggesting that breakthrough research starts with identifying crisp, quantifiable gaps. \textbf{Cross-Domain Synthesis is the second most common pattern} (687 papers, 18.0\%), where researchers import ideas from other fields and engineer compatibility layers, demonstrating that successful innovation often involves borrowing and adapting rather than inventing from scratch. \textbf{Representation Shift appears in 401 papers (10.5\%)}, where changing core primitives or abstractions simplifies the problem. Beyond individual patterns, successful research combines multiple strategies into repeatable ``innovation recipes.'' The most powerful combination pairs \textbf{Gap-Driven Reframing with Representation Shift} (318 occurrences), representing a ``Reframe + New Primitive'' strategy. The second combination, \textbf{Cross-Domain Synthesis with Representation Shift} (233 occurrences), embodies an ``Import + Adapt'' approach, while the third, \textbf{Gap-Driven Reframing with Cross-Domain Synthesis} (204 occurrences), reflects ``Diagnose + Borrow.'' These findings provide both a quantitative understanding of how scientific progress occurs and practical frameworks for generating novel research ideas. Critically, by capturing these patterns in structured format, our dataset enables the development of AI research agents that can learn from expert researchers' reasoning trajectories and apply these patterns to accelerate future discoveries.

Our source code is available at \url{https://github.com/AmberLJC/Sci-Reasoning} and the dataset is available at \url{https://huggingface.co/datasets/AmberLJC/Sci-Reasoning}.

\section{Related Work}
\label{sec:related}


\subsection{Research Lineage and Citation Analysis}

The study of scientific influence through citation analysis has evolved from simple counts and network structures~\cite{garfield1955citation,bornmann2008citation} to analyzing citation context and intent~\cite{hernandez2016survey,cohan2019structural,lahiri2023citeprompt,jantsch2025finecite}. Recent work has focused on identifying key predecessors: the progenitor index~\cite{jo2022see} identifies the single most influential prior work, while research lineage graphs~\cite{ghosal2021towards} identify significant citations where papers heavily rely on cited work. PST-Bench~\cite{zhang2024pst} formalized ``Publication Source Tracing'' as a task, exploring statistical, graph-based, and language model approaches with limited success.

Our work differs by identifying a \emph{set} of its key predecessors rather than a single progenitor, and focusing on the \emph{reasoning content}---the specific intellectual moves and synthesis strategies---rather than just influence patterns. Complementing work on contribution extraction~\cite{pramanick2025nature,lawrence2020argument}, we capture not individual contribution statements but the \emph{synthesis process} showing how contributions emerge from combining multiple prior works.

\subsection{Scientific Understanding and Reasoning}

Several datasets address scientific reasoning from different angles. PeerRead~\cite{kang2018dataset} provides peer reviews from major ML/NLP venues, demonstrating that review discourse offers valuable signals about research quality. GPQA-Diamond~\cite{rein2023gpqa} provides PhD-level science questions testing knowledge retrieval, while Polymathic AI datasets~\cite{cranmer2024polymathic} focus on domain-specific problem-solving. Research on narrative science~\cite{morgan2017narrative,green2017usefulness} studies how scientific progress is constructed and communicated through rhetorical and argumentative structures.

Our dataset occupies a unique niche: it captures the \emph{structured intellectual trajectories} behind breakthrough research---not just which papers influenced each advance, but specifically how ideas were combined, extended, or reframed. We leverage community-validated signals (oral/spotlight status) similar to PeerRead but focus on intellectual lineage and the reasoning that leads to high-quality papers, providing large-scale structured data for quantitative analysis of scientific narratives.

\subsection{AI-Assisted Research and Discovery}

Recent advances in LLMs have enabled significant progress in automating scientific research, from specialized tools to autonomous agents executing complete research workflows~\cite{kon2025curie, kon2025expbenchaiconductai, zheng2025automation}. Systems like LitLLM~\cite{agarwal2024litllm} use retrieval-augmented generation for literature review, neural search engines like Exa~\cite{exa2024search} enable semantic paper discovery, while Research Knowledge Graphs~\cite{zloch2025research} provide machine-actionable representations of research relations.

Our dataset provides crucial training and evaluation data for such systems. Understanding how high-quality research synthesizes prior work is essential for developing AI agents that generate valid scientific reasoning, with our structured synthesis traces offering explicit examples of reasoning patterns for research ideation and literature analysis.

\section{Methodology}
\label{sec:methodology}

\begin{figure*}[t]
    \centering
    \includegraphics[width=\textwidth]{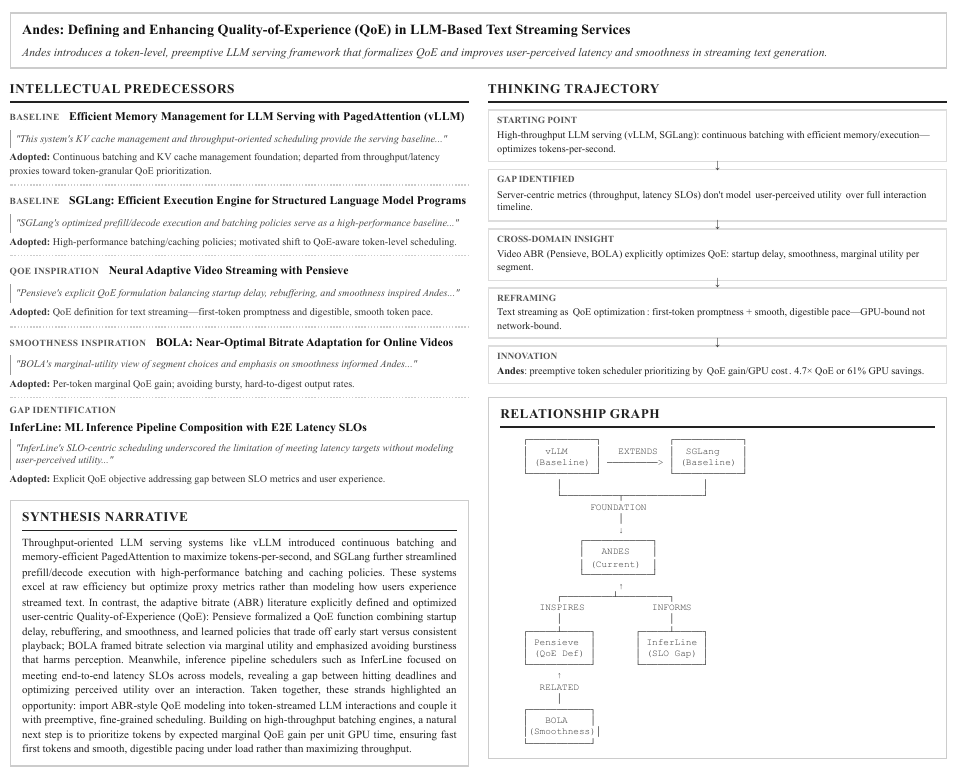}
    \caption{One complete dataset entry in Sci-Reasoning. For a target paper, our LLM pipeline identifies key predecessors (\S\ref{sec:lineage-tracing}) and generates structured metadata (predecessor role, relationship type) and rich synthesis narratives (\S\ref{sec:reasoning-link}). 
    See Appendix~\ref{app:example} for the complete JSON structure.}
    \label{fig:data-example}
    \end{figure*}

Our methodology for creating the Sci-Reasoning dataset consists of three phases: identifying high-quality papers, tracing their intellectual lineage, and articulating structured lineage graphs that capture the multi-dimensional relationships between papers (Figure~\ref{fig:data-example}). 

\subsection{High-Quality Paper Identification}
\label{sec:quality-identification}

Defining ``high quality'' is challenging and often subjective. We adopt a simple, defensible approach: a paper is considered high-quality if it was accepted for an \textbf{Oral or Spotlight presentation} at NeurIPS, ICML, or ICLR. This criterion uses the explicit judgment of conference program committees---comprising leading researchers in the field---as a direct proxy for a paper's significance and novelty at the time of publication.\footnote{Fully reproducible workflow: \url{https://www.orchestra-research.com/share/I0RLtZwO8Q2lUkFH}}

This approach offers several advantages: \textbf{(1) Community-validated:} The selection reflects collective expert assessment rather than individual opinion or post-hoc metrics. \textbf{(2) Signal strength:} Oral/Spotlight papers represent the top 1-5\% of submissions, providing a strong quality signal. \textbf{(3) Contemporaneous:} Unlike citation counts which accumulate over years, this signal is available immediately and reflects perceived quality at publication time. \textbf{(4) Reproducible:} The criterion is objective and easily replicated by other researchers.

We apply this criterion to Oral and Spotlight presentations from NeurIPS, ICML, and ICLR, (2023-2025). This results in 3,819 high-quality papers that serve as targets for lineage tracing. This scale allows us to capture diverse methodological approaches and intellectual trajectories across the major machine learning conferences while maintaining quality through our LLM-accelerated pipeline with human validation for quality assurance.

\subsection{Intellectual Lineage Tracing}
\label{sec:lineage-tracing}

For each high-quality target paper, we identify 5-10 key predecessors that form its intellectual foundation. This range reflects our observation that most breakthroughs synthesize insights from multiple sources rather than extending a single prior work.

We employ a single-pass LLM analysis using GPT-5 to process the full text of each target paper. The LLM is prompted to: (1) parse all internal citations and link them to the bibliography, (2) analyze the linguistic context surrounding each citation (e.g., ``building upon,'' ``inspired by,'' ``addresses the limitation of''), (3) count citation frequency across sections, (4) synthesize this evidence to rank cited works by importance, and (5) select the top 5-10 most significant predecessors that form the intellectual foundation of the target paper, ensuring diversity across contribution types (e.g., methodology, problem formulation, and baselines) rather than selecting only methodologically similar works. The complete prompt is provided in Appendix~\ref{app:prompts}.

This consolidated approach leverages LLMs' ability to perform holistic, end-to-end reasoning over long documents, reducing pipeline complexity compared to multi-step traditional NLP approaches~\cite{cohan2019structural}. The automated diversity-aware selection ensures systematic coverage of different predecessor roles while maintaining scalability across our dataset of 3,819 papers.

\subsection{Intellectual Connection Synthesis}
\label{sec:reasoning-link}

The core contribution of our methodology is representing intellectual lineage as structured ``Lineage Graphs'' that combine rich natural language narratives with queryable annotations. Each edge in the Lineage Graph connects a source (predecessor) paper to a target (current) paper, annotated across multiple dimensions\footnote{Fully reproducible workflow: \url{https://www.orchestra-research.com/share/oca4ADYe7gNmgh1T}}:

\textbf{Predecessor Role and Relationship Type:} Each connection is characterized along two complementary dimensions using LLM-based structured annotation. First, the model annotates the \emph{predecessor role}---what function the source paper serves in the intellectual lineage, such as providing methodological foundations, theoretical concepts, benchmarks for comparison, problem formulations, necessary resources, or inspiration from related domains. Second, the model captures the \emph{relationship type}---how the target paper builds upon the source, whether by extending, combining, bridging, addressing limitations of, or reframing ideas from the predecessor. Multiple relationship types may exist between the same paper pair.

\textbf{Synthesis Narrative:} An LLM-generated paragraph (200-400 words) explaining the intellectual synthesis that occurred. Each narrative follows a two-part structure: first, establishing the context by discussing the prior work and its key contributions relevant to the current paper; second, synthesizing how this prior work inspires and enables the target paper's contribution. This structure captures the ``story'' of how the target paper combined insights from its predecessors, identifying specific \emph{intellectual moves}, \emph{gaps identified}, and \emph{insight types} that enabled the scientific contribution. The unified predecessor identification and synthesis prompt (Appendix~\ref{app:prompts}) guides the LLM to identify key predecessors and reconstruct the authors' thinking trajectory as a narrative of intellectual discovery.

The complete schema is detailed in Appendix~\ref{app:handbook}. This structured representation makes the dataset queryable and analyzable at scale, while the natural language narratives provide the rich context needed to understand scientific reasoning.

\subsection{Quality Validation}
\label{sec:quality-validation}

Our methodology employs a fully automated LLM pipeline for scalability, with human validation to ensure quality.
A critical enabler of this approach is frontier LLMs' demonstrated capability to comprehend $\sim 15$ page papers and extract nuanced intellectual relationships---a task requiring human-level understanding of implicit reasoning and contextual synthesis that no existing automated method (citation analysis, topic modeling) systematically captures at scale.
We validate the pipeline using ~30 papers from domains we are familiar with, where we have ground-truth knowledge of predecessors and relationships. We iteratively refine all pipeline components---LLM prompts for candidate generation, diversity selection criteria, and structured annotation generation (Section~\ref{sec:reasoning-link})---until GPT-5 outputs consistently match ground truth. We select GPT-5 as our primary model based on an ablation study (Section~\ref{sec:model-ablation}) that demonstrates it achieves optimal cost-quality balance for predecessor extraction. Once validated, we apply the automated pipeline to all 3,819 papers.

For quality assurance at scale, we employ a two-tier validation strategy. For low-confidence cases (GPT-5 self-reported scores <0.7 in structured output), we employ multi-model cross-validation with Claude Opus 4.5 and Google Gemini 3.0. High agreement across models confirms reliability; disagreements trigger manual expert review. This human-in-the-loop approach concentrates expert attention on genuinely ambiguous cases (approximately 3\% of papers) while leveraging the complementary strengths of frontier LLMs for the majority of cases.

\section{Patterns Analysis of AI Research Innovation}
\label{sec:analysis}

\begin{figure}[t]
    \centering
    \includegraphics[width=\columnwidth]{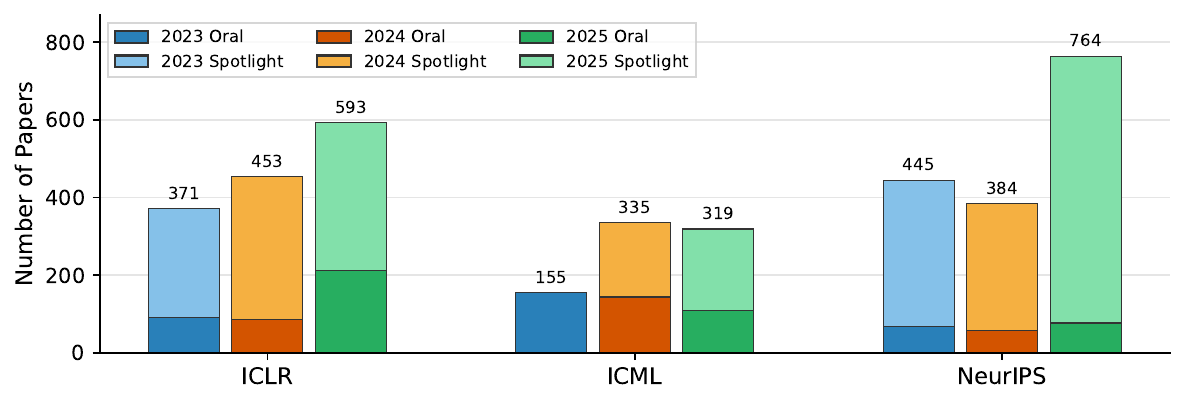}
    \caption{Distribution of papers in the Sci-Reasoning dataset across conferences, years, and presentation types (Oral and Spotlight).  }
    \label{fig:dataset-distribution}
\end{figure}

We analyze the Sci-Reasoning dataset (Figure~\ref{fig:dataset-distribution}; see Appendix~\ref{app:dataset-notes} for collection criteria) to uncover recurring patterns of innovation in high-quality AI research. By systematically examining the synthesis narratives of 3,819 Oral and Spotlight papers from NeurIPS, ICML, and ICLR (2023-2025), we identify 15 distinct \textbf{thinking patterns}---the cognitive strategies researchers employ to develop breakthrough ideas (complete descriptions in Appendix~\ref{app:complete-patterns}). These patterns represent not merely taxonomic categories, but actionable frameworks for systematic research ideation. Our analysis reveals how successful researchers diagnose gaps, reframe problems, synthesize cross-domain insights, and validate novel contributions.

\subsection{Thinking Pattern Taxonomy Derivation}
\label{sec:pattern-taxonomy}

To identify recurring cognitive strategies in breakthrough research, we develop a systematic taxonomy of thinking patterns through iterative LLM-based discovery and consolidation.\footnote{Fully reproducible workflow: \url{https://www.orchestra-research.com/share/BsrSD96SQyXjxZV7}}

\paragraph{Phase 1: Pattern Discovery.} We employ stratified sampling~\citep{patton2001qualitative} across conferences (NeurIPS, ICML, ICLR), years (2023--2025), and presentation types (Oral, Spotlight), selecting 10 batches of 35 papers each (350 total). For each batch, an LLM analyzes the synthesis narratives to identify recurring intellectual moves---the cognitive strategies authors use to develop contributions (detailed prompts in Appendix~\ref{app:prompts}). This batch-wise process yields approximately 190 raw patterns, intentionally allowing redundancy to capture patterns at different abstraction levels.

\paragraph{Phase 2: Taxonomy Consolidation.} An LLM consolidates the 190 raw patterns into 15 canonical thinking patterns by: (1) identifying semantically similar patterns across batches and (2) merging patterns at different abstraction levels (consolidation prompts in Appendix~\ref{app:prompts}). Each canonical pattern includes: a descriptive name, explanation of the cognitive move, and illustrative examples.

\paragraph{Phase 3: Full Classification and Analysis.} Using the 15-pattern taxonomy, we classify all 3,819 papers, processing synthesis narratives in batches of 5 to assign both a primary pattern (dominant strategy) and secondary pattern (supporting strategy, if present). The classified dataset enables systematic analysis of pattern distributions, temporal trends, conference-specific preferences, and pattern co-occurrence, identifying dominant patterns and successful pattern combinations.

\subsection{Dominant Thinking Patterns}
\label{sec:dominant-patterns}

\begin{figure}[t]
\centering
\includegraphics[width=\columnwidth]{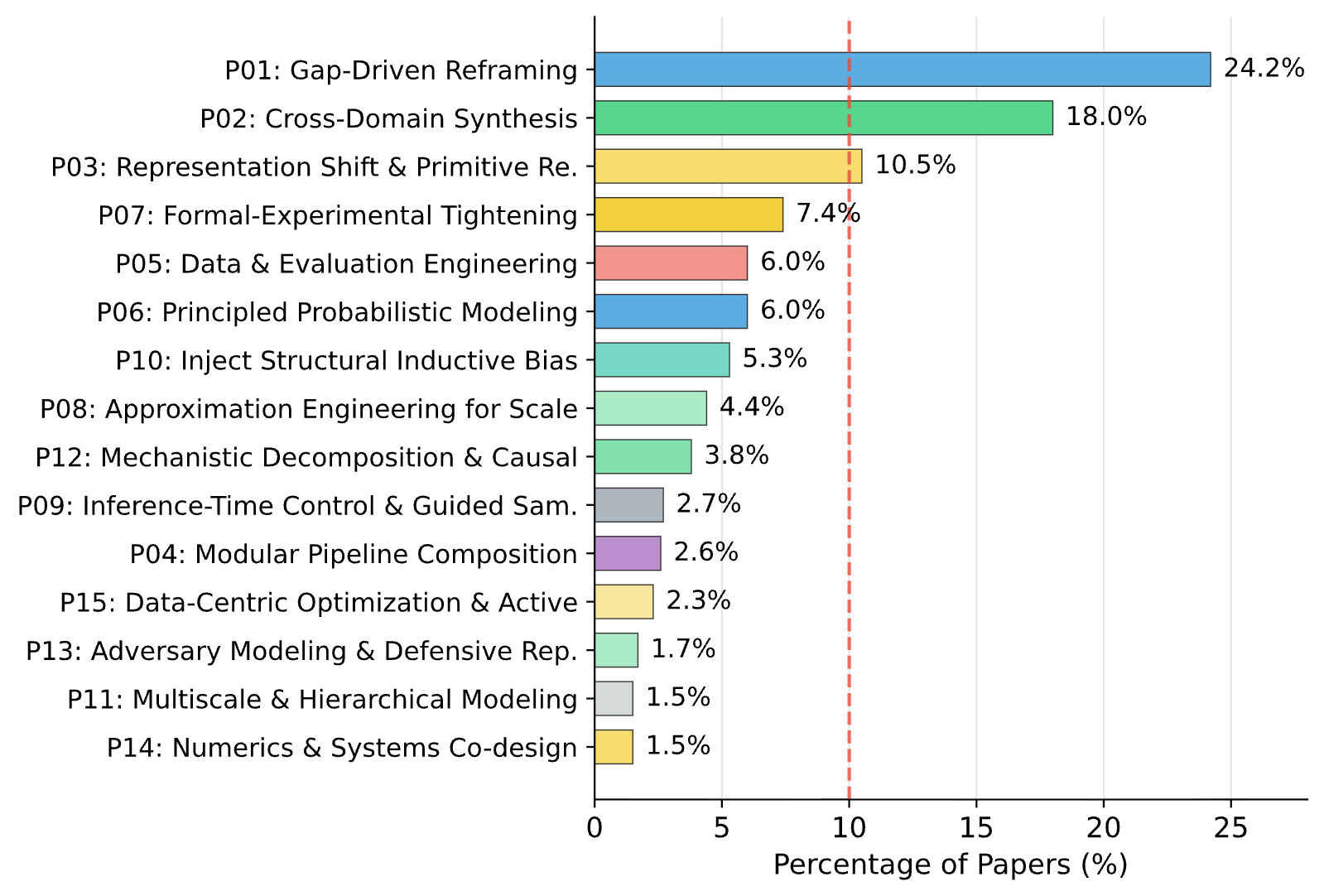}
\caption{Distribution of the 15 identified thinking patterns across 3,819 papers. The top three patterns---Gap-Driven Reframing, Cross-Domain Synthesis, and Representation Shift---account for 52.7\% of all papers.}
\label{fig:pattern-distribution}
\end{figure}

Three patterns dominate the landscape, accounting for 52.7\% of all analyzed papers, while the distribution follows a power law with a long tail of specialized strategies (Figure~\ref{fig:pattern-distribution}).

\begin{figure}[t]
    \centering
    \includegraphics[width=\columnwidth]{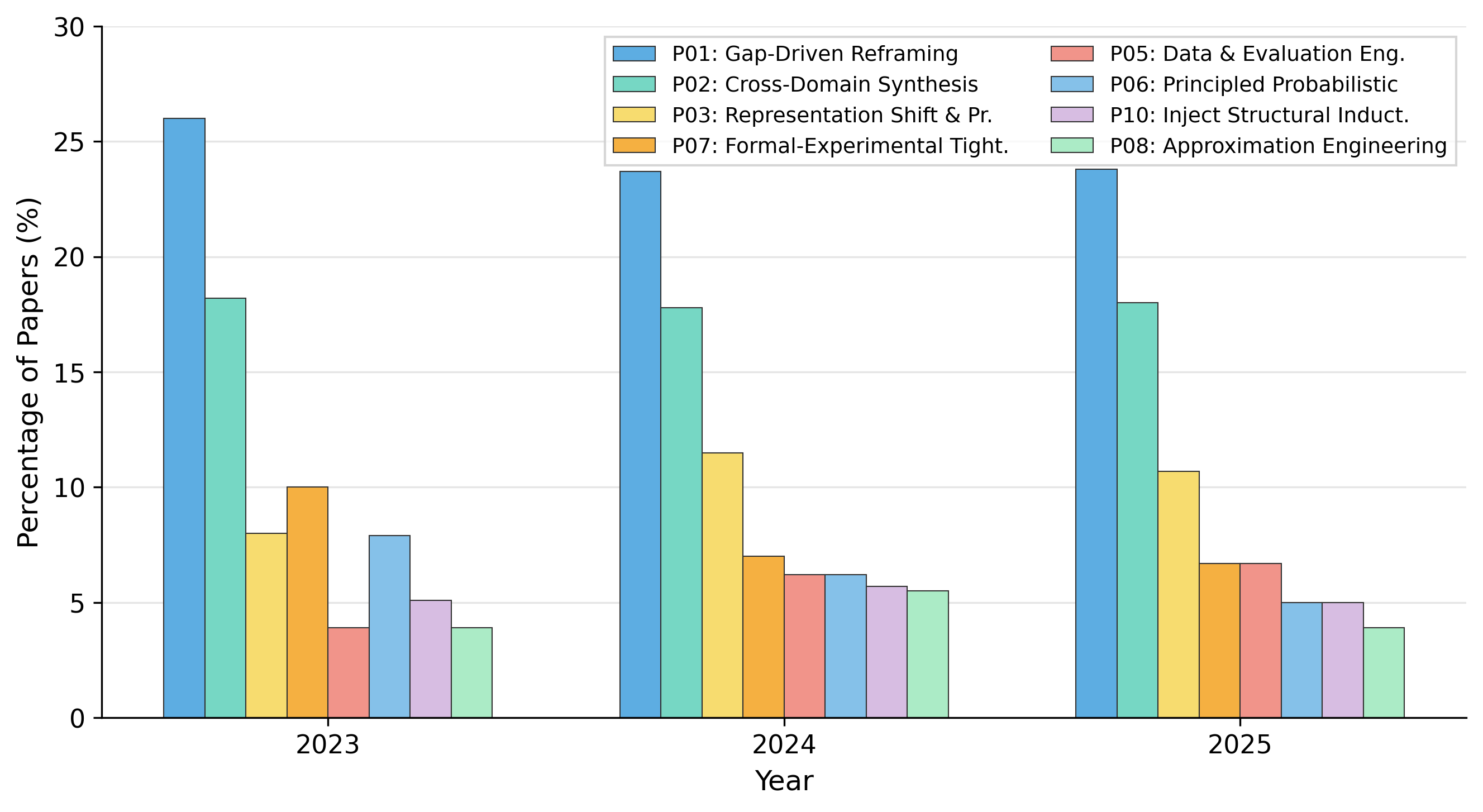}
    \caption{Temporal evolution of the top 5 thinking patterns from 2023 to 2025. While Gap-Driven Reframing remains stable, Representation Shift peaked in 2024, Formal-Experimental approaches are declining, and Data/Evaluation engineering is rising.}
    \label{fig:year-trends}
    \end{figure}

\textbf{Gap-Driven Reframing} (24.2\%) converts limitations into design constraints that guide solution design. Rather than treating failures as obstacles, researchers reframe them as specifications for better-suited approaches---e.g., reframing autoregressive image modeling from next-token to next-scale prediction transforms a scalability limitation into a principled architectural choice.

\textbf{Cross-Domain Synthesis} (18.0\%) transplants solutions from adjacent fields by identifying abstract constraints and engineering compatibility layers. Breakthroughs emerge from recognizing that analogous solutions exist elsewhere, such as fusing quantum circuits with transformer attention~\cite{born2025quantumdoublystochastictransformers} or importing control-theoretic stability into reinforcement learning.

\textbf{Representation Shift} (10.5\%) replaces a problem's fundamental primitives---pixels, tokens, meshes---with alternatives that simplify inference or better capture constraints, such as replacing explicit meshes with neural implicit functions for 3D reconstruction. Together, these patterns reveal a consistent meta-strategy: reframe problems, import cross-domain solutions, or reconceptualize representations (detailed case studies in Appendix~\ref{app:cases}).

\subsection{Temporal Evolution of Patterns}
\label{sec:temporal-evolution}

Temporal trends reveal that problem diagnosis and reformulation remain the fundamental engine of innovation (Figure~\ref{fig:year-trends}). The decline in Formal-Experimental Tightening suggests theoretical analysis is becoming a supporting element rather than a primary one, while growth in Data/Evaluation Engineering reflects the field's maturation toward rigorous empirical methodology.

\subsection{Conference-Specific Patterns}
\label{sec:conference-patterns}

\begin{figure}[t]
\centering
\includegraphics[width=\columnwidth]{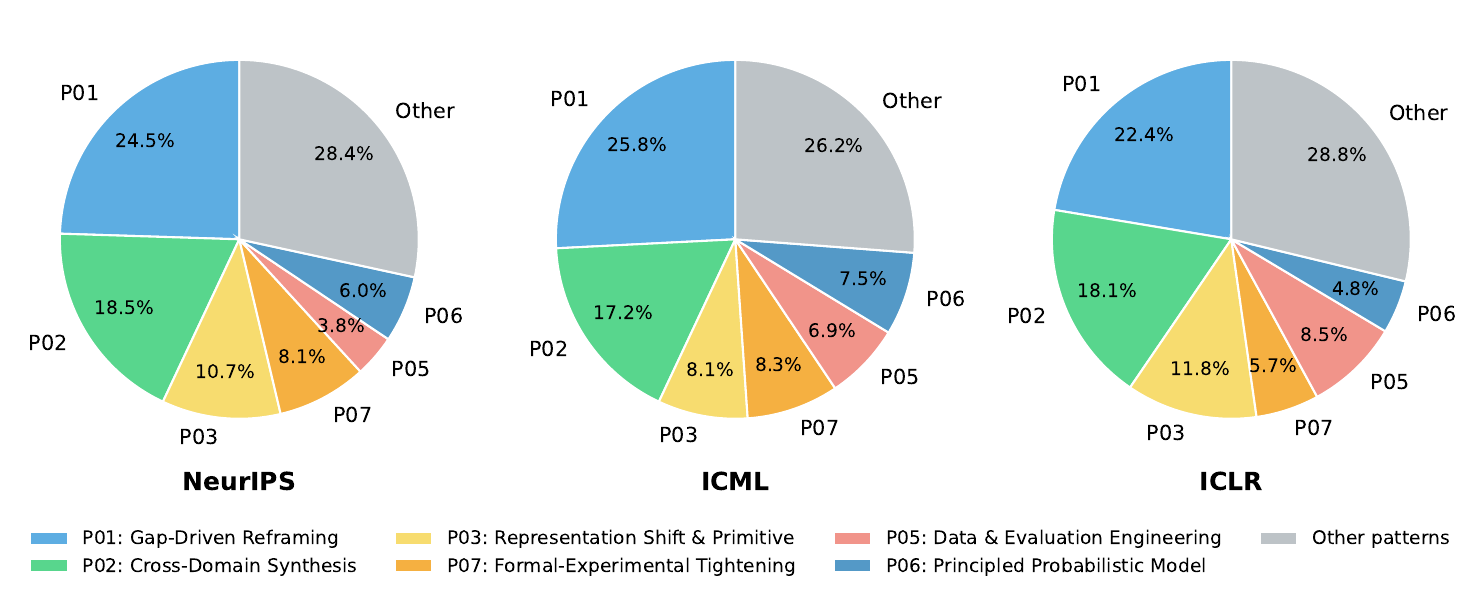}
\caption{Conference-specific distribution of thinking patterns. ICML shows stronger preference for formal methods (8.3\%) and probabilistic modeling (7.5\%), ICLR emphasizes representation innovation (11.8\%) and benchmarking (8.5\%), while NeurIPS maintains balanced, cross-disciplinary coverage.}
\label{fig:conference-comparison}
\end{figure}

Beyond their shared emphasis on Gap-Driven Reframing, the three major conferences exhibit distinct intellectual cultures (Figure~\ref{fig:conference-comparison}). These differences suggest tailoring submission strategies: ICML submissions benefit from mathematical rigor and theoretical guarantees, ICLR submissions should highlight architectural innovations and empirical methodology, while NeurIPS favors broad applicability and cross-disciplinary synthesis.

\subsection{Research Pattern Combinations}
\label{sec:pattern-combinations}

\begin{figure}[t]
    \centering
    \includegraphics[width=\columnwidth]{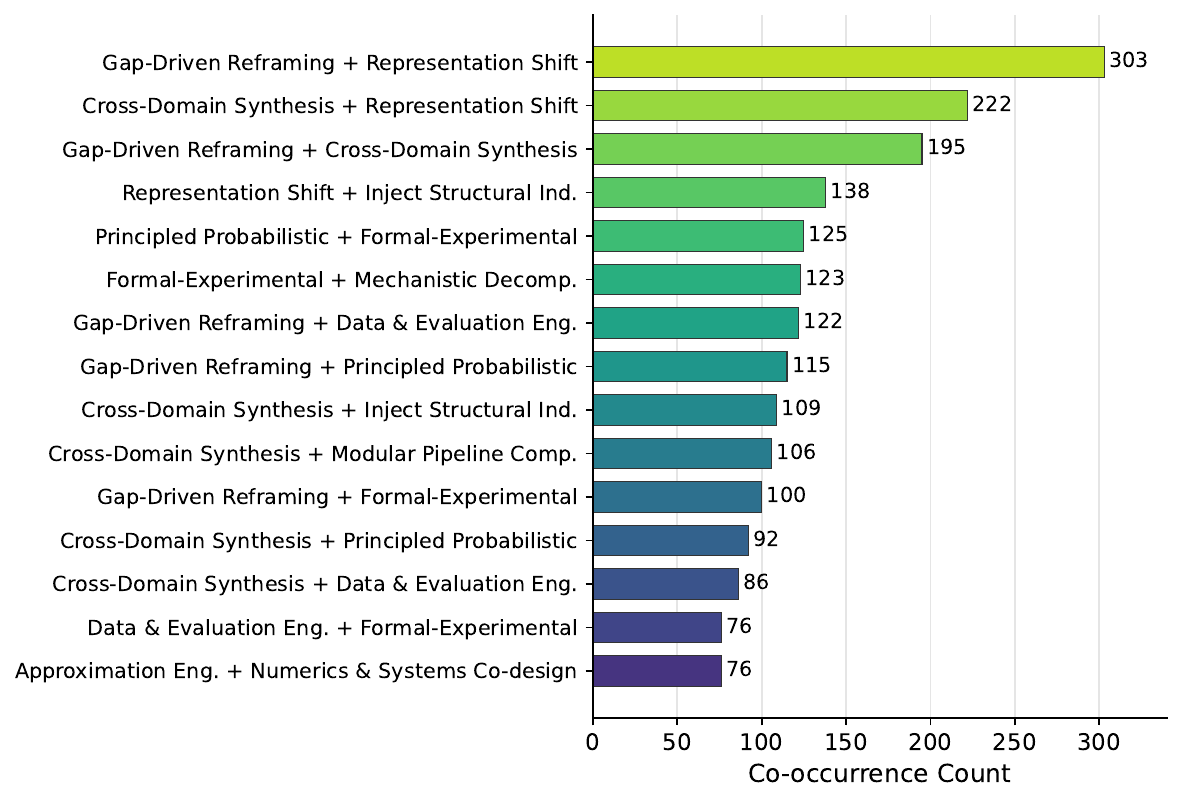}
    \caption{Top 10 pattern combinations (primary + secondary). The most successful research employs multiple thinking patterns: ``Reframe + New Primitive'' (318), ``Import + Adapt'' (233), and ``Diagnose + Borrow'' (204) represent repeatable innovation recipes.}
    \label{fig:pattern-pairs}
    \end{figure}

Research breakthroughs rarely employ a single thinking pattern in isolation. Analysis of secondary patterns reveals systematic combinations that function as repeatable ``research recipes'' (Figure~\ref{fig:pattern-pairs}). The most frequent combination pairs Gap-Driven Reframing with Representation Shift (318 occurrences), embodying a powerful two-step strategy: diagnose a limitation, then introduce a new primitive that sidesteps it. This ``Reframe + New Primitive'' recipe transforms conceptual insights into concrete architectural innovations.

The second most common combination, Cross-Domain Synthesis with Representation Shift (233 occurrences), represents an ``Import + Adapt'' strategy: borrow a mechanism from another field and modify its representation to fit the target domain. 
The third combination, Gap-Driven Reframing with Cross-Domain Synthesis (204 occurrences), embodies ``Diagnose + Borrow'': identify what's missing, then search for solutions in adjacent fields.
 
These patterns suggest that breakthrough research follows a meta-pattern: \emph{Diagnose} a gap, \emph{Represent} it differently or \emph{Import} from elsewhere, then \emph{Validate} rigorously. For detailed statistical analyses for all above analyses and practical guidance, see Appendix~\ref{app:stats}. 

Finally, we conduct an assessment of whether LLMs can predict research directions from intellectual predecessors (Section~\ref{sec:experiments}). Testing four frontier models on 77 NeurIPS 2025 Oral papers, we find that Gemini 2.5 Pro achieves 49.35\% Hit@10 accuracy, demonstrating that our methodology captures meaningful intellectual relationships while also revealing a natural predictability ceiling that validates genuine research creativity.

\section{Evaluation}
\label{sec:experiments}

\subsection{Evaluating LLM on Research Ideation}

We evaluate whether LLMs can generate research ideas matching real publications given only their intellectual predecessors.\footnote{Fully reproducible workflow: \url{https://www.orchestra-research.com/share/xkueih5uZdNu4Lb8}} Our pipeline takes intellectual predecessor papers from each NeurIPS 2025 Oral paper, retrieves the prior work paper content using Exa AI~\cite{exa2024search}, prompts an LLM to generate $k=10$ candidate research ideas, and evaluates semantic similarity between generated ideas and the actual published paper using an LLM judge.

We employ GPT-5.2 as an evaluator~\cite{zheng2023judging} to assess whether each generated idea semantically matches the ground truth paper. 
For each paper, the judge receives both the generated ideas and the actual paper's title and contribution statement, then determines if they represent similar research directions by considering: (1) whether they address the same core problem, (2) whether they propose similar methodological approaches, and (3) whether the generated idea, if fully developed, would result in a similar contribution. The judge returns a structured JSON response containing a binary match decision, a confidence score (0-1), and a brief reasoning explanation. The complete evaluation prompt is provided in Appendix~\ref{app:llm-judge-prompt}. We measure success using Hit@10: whether any of the 10 generated ideas matches the ground truth paper according to the LLM judge. We evaluate on 77 NeurIPS 2025 Oral papers, using four frontier models: GPT-5.2, Claude Sonnet 4, Claude Opus 4, and Gemini 2.5 Pro.

Table~\ref{tab:hit_rates} presents our main findings. Gemini 2.5 Pro achieves the highest Hit@10 rate of 49.35\%, successfully predicting nearly half of research directions from intellectual predecessors alone. 
The 19.5 percentage point performance gap demonstrates meaningful differences in research direction prediction capabilities across frontier models.
Research ideation exhibits a many-to-many relationship: the same intellectual predecessors can inspire multiple valid research directions, while different predecessor combinations may lead to similar ideas. This means our Hit@10 metric likely provides a conservative estimate of model capabilities, as generated ideas that don't match the published paper may still represent valid research directions.

\begin{table}[t]
\centering
\small
\begin{tabular}{lc}
\toprule
\textbf{Model} & \textbf{Hit@10 (\%)} \\
\midrule
Gemini 2.5 Pro & 49.35 \\
Claude Opus 4 & 42.86 \\
GPT-5.2 & 38.89 \\
Claude Sonnet 4 & 29.87 \\
\bottomrule
\end{tabular}
\caption{Hit@10 rates on predicting research directions.}
\label{tab:hit_rates}
\end{table}

\begin{table}[t]
\centering
\small
\begin{tabular}{lc}
\toprule
\textbf{Model} & \textbf{Recall} \\
\midrule
GPT-5 & \textbf{89.73\%} \\
GPT-5.2 & 87.47\% \\
GPT-4.1 & 78.00\% \\
GPT-5-mini & 68.53\% \\
\bottomrule
\end{tabular}
\caption{Model ablation for predecessor extraction.}
\label{tab:model-ablation}
\end{table}

\subsection{Evaluating LLM on Predecessor Extraction}
\label{sec:model-ablation}

To ensure our pipeline balances cost and quality, we conduct an ablation study evaluating four OpenAI models on the predecessor extraction task using a validation set of 77 Oral/Spotlight papers with ground-truth predecessors.

Table~\ref{tab:model-ablation} shows that GPT-5 achieves the highest recall, outperforming even the newer GPT-5.2. This surprising result suggests that model generation alone does not guarantee better performance on specialized academic tasks. GPT-4.1 shows consistent performance with zero complete failures but exhibits a substantial quality gap. GPT-5-mini demonstrates the largest capability gap with frequent failures on specialized theoretical papers, indicating insufficient domain knowledge for academic tasks.
Based on these results, we select GPT-5 as our primary model for dataset construction, achieving the optimal cost-quality balance with superior performance and lower API costs.  

\section{Limitations}
\label{sec:limitations}

We acknowledge several important limitations of our work that should be considered when interpreting results and planning future research.

\textbf{Logic of Justification vs. Discovery:} We analyze the \emph{logic of justification}---how researchers present contributions in published papers---rather than the \emph{logic of discovery}---the actual thought process behind breakthroughs~\cite{reichenbach1938experience}. Published papers are polished narratives that may omit failed experiments, abandoned hypotheses, serendipity, or external influences. Our dataset captures researchers' final presentations, which may differ from the chronological reality of how insights emerged. This inherent constraint means only successful reasoning paths can be systematically analyzed; we make this explicit to avoid overstating claims about understanding creativity.

\textbf{Temporal Constraint:} Our dataset represents a snapshot of research patterns in 2023-2025. Innovation patterns may evolve as fields mature, new methodologies emerge, or community norms change. Longitudinal analysis tracking how patterns shift over decades would be valuable but is beyond our current scope.

\textbf{Conference and Selection Criteria Scope:} Restricting our dataset to Oral/Spotlight presentations at NeurIPS, ICML, and ICLR may favor empirical breakthroughs over theoretical work, reflecting conference-specific biases rather than long-term impact. This focus misses other AI venues (AAAI, IJCAI, UAI), interdisciplinary research (computational biology, neuroscience), and non-ML AI areas (symbolic AI, knowledge representation). Our innovation patterns may be specific to mainstream ML circa 2023-2025 and may not generalize to other domains or AI subfields.


Despite these limitations, we believe Sci-Reasoning makes a valuable contribution by providing the first large-scale, structured dataset for studying scientific reasoning in AI research, demonstrating its reliability through quality evaluation, and revealing actionable patterns of innovation.

\section{Conclusion}
\label{sec:conclusion}

We present Sci-Reasoning, a dataset capturing structured intellectual lineage behind scientific breakthroughs in top-tier AI research. By transforming implicit reasoning into queryable lineage graphs, our work reveals systematic patterns in how researchers diagnose gaps, synthesize cross-domain insights, and combine ideas into novel contributions. These patterns provide both scientific understanding of innovation mechanisms and practical frameworks for AI research agents. By making scientific reasoning explicit, our work advances the augmentation of scientific discovery.

\clearpage
\bibliography{references}

@inproceedings{cohan2019structural,
  title={Structural scaffolds for citation intent classification in scientific publications},
  author={Cohan, Arman and Ammar, Waleed and Van Zuylen, Madeleine and Cady, Field},
  booktitle={Proceedings of NAACL},
  pages={3586--3596},
  year={2019}
}

@inproceedings{kang2018dataset,
  title={A dataset of peer reviews (PeerRead): Collection, insights and NLP applications},
  author={Kang, Dongyeop and Ammar, Waleed and Dalvi, Bhavana and van Zuylen, Madeleine and Kohlmeier, Sebastian and Hovy, Eduard and Schwartz, Roy},
  booktitle={Proceedings of NAACL-HLT},
  pages={1647--1661},
  year={2018}
}

@inproceedings{lawrence2020argument,
  title={Argument mining: A survey},
  author={Lawrence, John and Reed, Chris},
  booktitle={Computational Linguistics},
  volume={46},
  number={4},
  pages={765--818},
  year={2020}
}

@article{zheng2023judging,
  title={Judging LLM-as-a-judge with MT-Bench and Chatbot Arena},
  author={Zheng, Lianmin and Chiang, Wei-Lin and Sheng, Ying and Zhuang, Siyuan and Wu, Zhanghao and Zhuang, Yonghao and Lin, Zi and Li, Zhuohan and Li, Dacheng and Xing, Eric P and others},
  journal={arXiv preprint arXiv:2306.05685},
  year={2023}
}

@article{bornmann2008citation,
  title={What do citation counts measure? A review of studies on citing behavior},
  author={Bornmann, Lutz and Daniel, Hans-Dieter},
  journal={Journal of Documentation},
  volume={64},
  number={1},
  pages={45--80},
  year={2008},
  publisher={Emerald Group Publishing Limited}
}

@misc{kon2025expbenchaiconductai,
      title={EXP-Bench: Can AI Conduct AI Research Experiments?}, 
      author={Patrick Tser Jern Kon and Jiachen Liu and Xinyi Zhu and Qiuyi Ding and Jingjia Peng and Jiarong Xing and Yibo Huang and Yiming Qiu and Jayanth Srinivasa and Myungjin Lee and Mosharaf Chowdhury and Matei Zaharia and Ang Chen},
      year={2025},
      eprint={2505.24785},
      archivePrefix={arXiv},
      primaryClass={cs.AI},
      url={https://arxiv.org/abs/2505.24785}, 
}

@article{zhang2024pst,
  title={PST-Bench: Tracing and Benchmarking the Source of Publications},
  author={Zhang, Fanjin and Cao, Kun and Cen, Yukuo and Yu, Jifan and Yin, Da and Tang, Jie},
  journal={arXiv preprint arXiv:2402.16009},
  year={2024}
}

@inproceedings{pramanick2025nature,
  title={The Nature of NLP: Analyzing Contributions in NLP Papers},
  author={Pramanick, Aniket and Hou, Yufang and Mohammad, Saif M. and Gurevych, Iryna},
  booktitle={Proceedings of the 63rd Annual Meeting of the Association for Computational Linguistics (Volume 1: Long Papers)},
  pages={25169--25191},
  year={2025},
  address={Vienna, Austria}
}

@inproceedings{lahiri2023citeprompt,
  title={CitePrompt: Using Prompts to Identify Citation Intent in Scientific Papers},
  author={Lahiri, Avishek and Sanyal, Debarshi Kumar and Mukherjee, Imon},
  booktitle={Proceedings of the 2023 ACM/IEEE Joint Conference on Digital Libraries},
  pages={51--55},
  year={2023},
  organization={ACM}
}

@inproceedings{jantsch2025finecite,
  title={FineCite: A Novel Approach For Fine-Grained Citation Context Analysis},
  author={Jantsch, Lasse M. and Koh, Dong-Jae and Yoon, Seonghwan and Lee, Jisu and Lauscher, Anne and Suh, Young-Kyoon},
  booktitle={Findings of the Association for Computational Linguistics: ACL 2025},
  pages={24525--24542},
  year={2025},
  address={Vienna, Austria}
}

@article{hernandez2016survey,
  title={Survey about citation context analysis: Tasks, techniques, and resources},
  author={Hernández-Álvarez, Myriam and Gómez, José Manuel},
  journal={Natural Language Engineering},
  volume={22},
  number={3},
  pages={327--349},
  year={2016},
  publisher={Cambridge University Press}
}

@book{patton2001qualitative,
  title={Qualitative research and evaluation methods},
  author={Patton, Michael Quinn},
  edition={3},
  year={2001},
  publisher={Sage Publications},
  address={Thousand Oaks, CA}
}

@article{ouyang2022training,
  title={Training language models to follow instructions with human feedback},
  author={Ouyang, Long and Wu, Jeffrey and Jiang, Xu and Almeida, Diogo and Wainwright, Carroll and Mishkin, Pamela and Zhang, Chong and Agarwal, Sandhini and Slama, Katarina and Ray, Alex and others},
  journal={Advances in Neural Information Processing Systems},
  volume={35},
  pages={27730--27744},
  year={2022}
}

@book{reichenbach1938experience,
  title={Experience and prediction: An analysis of the foundations and the structure of knowledge},
  author={Reichenbach, Hans},
  year={1938},
  publisher={University of Chicago Press}
}

@article{garfield1955citation,
  title={Citation indexes for science},
  author={Garfield, Eugene},
  journal={Science},
  volume={122},
  number={3159},
  pages={108--111},
  year={1955},
  publisher={American Association for the Advancement of Science}
}

@article{jo2022see,
  title={See further upon the giants: Quantifying intellectual lineage in science},
  author={Jo, Woo Seong and Liu, Lu and Wang, Dashun},
  journal={Quantitative Science Studies},
  volume={3},
  number={2},
  pages={319--330},
  year={2022},
  publisher={MIT Press}
}

@article{ghosal2021towards,
  title={Towards establishing a research lineage via identification of significant citations},
  author={Ghosal, Tirthankar and Edithal, Vignesh and Ekbal, Asif and Bhattacharyya, Pushpak and Tsatsaronis, George and Verma, Sriparna},
  journal={Quantitative Science Studies},
  volume={2},
  number={4},
  pages={1511--1542},
  year={2021},
  publisher={MIT Press}
}

@article{vaswani2017attention,
  title={Attention is all you need},
  author={Vaswani, Ashish and Shazeer, Noam and Parmar, Niki and Uszkoreit, Jakob and Jones, Llion and Gomez, Aidan N and Kaiser, {\L}ukasz and Polosukhin, Illia},
  journal={Advances in Neural Information Processing Systems},
  volume={30},
  year={2017}
}

@misc{born2025quantumdoublystochastictransformers,
  title={Quantum Doubly Stochastic Transformers},
  author={Born, Jannis and Skogh, Filip and Rhrissorrakrai, Kahn and Utro, Filippo and Wagner, Nico and Sobczyk, Aleksandros},
  year={2025},
  eprint={2504.16275},
  archivePrefix={arXiv},
  primaryClass={cs.LG},
  url={https://arxiv.org/abs/2504.16275}
}

@article{ramesh2022hierarchical,
  title={Hierarchical text-conditional image generation with clip latents},
  author={Ramesh, Aditya and Dhariwal, Prafulla and Nichol, Alex and Chu, Casey and Chen, Mark},
  journal={arXiv preprint arXiv:2204.06125},
  year={2022}
}

@article{rein2023gpqa,
  title={GPQA: A graduate-level google-proof Q\&A benchmark},
  author={Rein, David and Hou, Betty Li and Stickland, Asa Cooper and Petty, Jackson and Pang, Richard Yuanzhe and Dirani, Julien and Michael, Julian and Bowman, Samuel R},
  journal={arXiv preprint arXiv:2311.12022},
  year={2023}
}

@article{cranmer2024polymathic,
  title={The Polymathic AI Project: Laying the foundations for a new era of scientific machine learning},
  author={Cranmer, Miles and others},
  journal={Machine Learning: Science and Technology},
  volume={5},
  number={4},
  pages={045043},
  year={2024},
  publisher={IOP Publishing}
}

@incollection{morgan2017narrative,
  title={Narrative science},
  author={Morgan, Mary S and Wise, M Norton},
  booktitle={The Cambridge History of Science: Volume 7, The Modern Social Sciences},
  pages={605--620},
  year={2017},
  publisher={Cambridge University Press}
}

@article{green2017usefulness,
  title={On the usefulness of narratives: An interdisciplinary review and theoretical model},
  author={Green, Melanie C and Donahue, Jay K},
  journal={Annals of the International Communication Association},
  volume={41},
  number={3-4},
  pages={227--245},
  year={2017},
  publisher={Taylor \& Francis}
}

@misc{liu2024andesdefiningenhancingqualityofexperience,
  title={Andes: Defining and Enhancing Quality-of-Experience in LLM-Based Text Streaming Services},
  author={Jiachen Liu and Jae-Won Chung and Zhiyu Wu and Fan Lai and Myungjin Lee and Mosharaf Chowdhury},
  year={2024},
  eprint={2404.16283},
  archivePrefix={arXiv},
  primaryClass={cs.DC},
  url={https://arxiv.org/abs/2404.16283}
}

@misc{exa2024search,
  title={Exa: Neural Search for the Age of AI},
  author={{Exa AI}},
  year={2024},
  howpublished={\url{https://exa.ai}}
}

@article{agarwal2024litllm,
  title={LitLLM: A Toolkit for Scientific Literature Review},
  author={Agarwal, Shubham and Sahu, Gaurav and Puri, Abhay and Laradji, Issam H. and Dvijotham, Krishnamurthy DJ and Stanley, Jason and Charlin, Laurent and Pal, Christopher},
  journal={arXiv preprint arXiv:2402.01788},
  year={2024}
}

@inproceedings{zheng2025automation,
  title={From Automation to Autonomy: A Survey on Large Language Models in Scientific Discovery},
  author={Zheng, Tianshi and Deng, Zheye and Tsang, Hong Ting and Wang, Weiqi and Bai, Jiaxin and Wang, Zihao and Song, Yangqiu},
  booktitle={Proceedings of the 2025 Conference on Empirical Methods in Natural Language Processing},
  pages={17744--17761},
  year={2025},
  address={Suzhou, China}
}

@misc{kon2025curie,
  title={Curie: Toward Rigorous and Automated Scientific Experimentation with AI Agents},
  author={Kon, Patrick Tser Jern and Liu, Jiachen and Ding, Qiuyi and Qiu, Yiming and Yang, Zhenning and Huang, Yibo and Srinivasa, Jayanth and Lee, Myungjin and Chowdhury, Mosharaf and Chen, Ang},
  year={2025},
  eprint={2502.16069},
  archivePrefix={arXiv},
  primaryClass={cs.AI},
  howpublished={\url{https://arxiv.org/abs/2502.16069}}
}

@inproceedings{zloch2025research,
  title={Research Knowledge Graphs: the Shifting Paradigm of Scholarly Information Representation},
  author={Zloch, Matthäus and Dessì, Danilo and D'Souza, Jennifer and Castro, Leyla Jael and Zapilko, Benjamin and Karmakar, Saurav and Mathiak, Brigitte and Stocker, Markus and Otto, Wolfgang and Auer, Sören and Dietze, Stefan},
  booktitle={Proceedings of the Extended Semantic Web Conference},
  year={2025}
}

\appendix
\clearpage

\section{Complete Dataset Example}
\label{app:example}

This section provides a complete example of a dataset entry, showing the full annotation for a single target paper with all intellectual predecessors, thinking trajectory, synthesis narrative, and relationship graph. The example uses the Andes paper~\cite{liu2024andesdefiningenhancingqualityofexperience}, which introduces a token-level, preemptive LLM serving framework that formalizes QoE and improves user-perceived latency and smoothness in streaming text generation.

\subsection*{Target Paper}

\textbf{Andes: Defining and Enhancing Quality-of-Experience (QoE) in LLM-Based Text Streaming Services}

\emph{Andes introduces a token-level, preemptive LLM serving framework that formalizes QoE and improves user-perceived latency and smoothness in streaming text generation.}

\subsection*{Intellectual Predecessors}

\paragraph{BASELINE: Efficient Memory Management for LLM Serving with PagedAttention (vLLM)}

\emph{``This system's KV cache management and throughput-oriented scheduling provide the serving baseline...''}

\textbf{Adopted:} Continuous batching and KV cache management foundation; departed from throughput/latency proxies toward token-granular QoE prioritization.

\paragraph{BASELINE: SGLang: Efficient Execution Engine for Structured Language Model Programs}

\emph{``SGLang's optimized prefill/decode execution and batching policies serve as a high-performance baseline...''}

\textbf{Adopted:} High-performance batching/caching policies; motivated shift to QoE-aware token-level scheduling.

\paragraph{QOE INSPIRATION: Neural Adaptive Video Streaming with Pensieve}

\emph{``Pensieve's explicit QoE formulation balancing startup delay, rebuffering, and smoothness inspired Andes...''}

\textbf{Adopted:} QoE definition for text streaming---first-token promptness and digestible, smooth token pace.

\paragraph{SMOOTHNESS INSPIRATION: BOLA: Near-Optimal Bitrate Adaptation for Online Videos}

\emph{``BOLA's marginal-utility view of segment choices and emphasis on smoothness informed Andes...''}

\textbf{Adopted:} Per-token marginal QoE gain; avoiding bursty, hard-to-digest output rates.

\paragraph{GAP IDENTIFICATION: InferLine: ML Inference Pipeline Composition with E2E Latency SLOs}

\emph{``InferLine's SLO-centric scheduling underscored the limitation of meeting latency targets without modeling user-perceived utility...''}

\textbf{Adopted:} Explicit QoE objective addressing gap between SLO metrics and user experience.

\subsection*{Thinking Trajectory}

\begin{description}
\item[STARTING POINT] High-throughput LLM serving (vLLM, SGLang): continuous batching with efficient memory/execution---optimizes tokens-per-second, not user timeline.
\item[GAP IDENTIFIED] Server-centric metrics (throughput, latency SLOs) don't model user-perceived utility over full interaction timeline.
\item[CROSS-DOMAIN INSIGHT] Video ABR (Pensieve, BOLA) explicitly optimizes QoE: startup delay, smoothness, marginal utility per segment.
\item[REFRAMING] Text streaming as QoE optimization: first-token promptness + smooth, digestible pace---GPU-bound not network-bound.
\item[INNOVATION] \textbf{Andes:} preemptive token scheduler prioritizing by QoE gain/GPU cost. 4.7× QoE or 61\% GPU savings.
\end{description}

\subsection*{Synthesis Narrative}

Throughput-oriented LLM serving systems like vLLM introduced continuous batching and memory-efficient PagedAttention to maximize tokens-per-second, and SGLang further streamlined prefill/decode execution with high-performance batching and caching policies. These systems excel at raw efficiency but optimize proxy metrics rather than modeling how users experience streamed text. In contrast, the adaptive bitrate (ABR) literature explicitly defined and optimized user-centric Quality-of-Experience (QoE): Pensieve formalized a QoE function combining startup delay, rebuffering, and smoothness, and learned policies that trade off early start versus consistent playback. BOLA framed bitrate selection via marginal utility and emphasized avoiding burstiness that harms perception. Meanwhile, inference pipeline schedulers such as InferLine focused on meeting end-to-end latency SLOs across models, revealing a gap between hitting deadlines and optimizing perceived utility over an interaction. Taken together, these strands highlighted an opportunity: import ABR-style QoE modeling into token-streamed LLM interactions and couple it with preemptive, fine-grained scheduling. Building on high-throughput batching engines, a natural next step is to prioritize tokens by expected marginal QoE gain per unit GPU time, ensuring fast first tokens and smooth, digestible pacing under load rather than maximizing throughput.

\subsection*{Relationship Graph}

The relationship graph illustrates how the intellectual predecessors combine to form the innovation:

\begin{itemize}
\item \textbf{vLLM (Baseline)} and \textbf{SGLang (Baseline)} \textrightarrow\ EXTENDS \textrightarrow\ \textbf{ANDES (Current)}: Foundation of efficient batching and execution.
\item \textbf{Pensieve (QoE Def)} \textrightarrow\ INSPIRES \textrightarrow\ \textbf{ANDES}: QoE formulation for text streaming.
\item \textbf{InferLine (SLO Gap)} \textrightarrow\ INFORMS \textrightarrow\ \textbf{ANDES}: Highlights gap between SLO targets and user experience.
\item \textbf{BOLA (Smoothness)} \textrightarrow\ RELATED \textrightarrow\ \textbf{Pensieve}: Both contribute to QoE modeling in video streaming.
\end{itemize}

\section{LLM Pipeline Prompts}
\label{app:prompts}

This section provides the detailed system prompts used in our LLM pipeline. We use a unified prompt that combines predecessor identification with synthesis narrative generation, ensuring that the identified predecessors and their relationships are coherently integrated into the intellectual lineage story. These prompts are incorporated into our automated pipeline using GPT-5.

\subsection{Unified Predecessor Identification and Synthesis Prompt}

\textbf{Purpose:} Identify key intellectual predecessors and generate synthesis narratives that explain how they collectively inspired the target paper's innovation.

\textbf{System Prompt:}

\begin{quote}
You are an expert AI research analyst. Your task is to identify the KEY PRIOR WORKS that DIRECTLY led to a research paper's core innovation.

\paragraph{CRITICAL: Focus on DIRECT Intellectual Lineage}

You must identify papers that are \textbf{directly responsible} for the current paper's main contributions. Ask yourself:
\begin{itemize}
\item ``Without this prior work, would the current paper's core idea exist?''
\item ``Did this prior work directly inspire, enable, or motivate the KEY INNOVATION?''
\item ``Is this paper cited in the Introduction or Related Work as a PRIMARY influence?''
\end{itemize}

\paragraph{DO NOT INCLUDE:}
\begin{itemize}
\item Generic infrastructure/tools (e.g., PyTorch, CUDA, standard attention mechanisms)
\item Complementary optimizations that are orthogonal to the main contribution
\item Papers that share the same domain but don't directly influence the core idea
\item Standard baselines that are just compared against without deeper connection
\item Well-known foundational works that everyone cites but aren't specific to this innovation
\end{itemize}

\paragraph{DO INCLUDE:}
\begin{itemize}
\item Papers whose specific IDEAS, METHODS, or FINDINGS directly shaped the current work
\item Papers whose LIMITATIONS or GAPS the current paper explicitly addresses
\item Papers that introduced the PROBLEM FORMULATION the current paper builds on
\item Papers whose TECHNIQUES are directly extended or modified
\item Papers that provide the KEY INSIGHT that the current paper leverages
\end{itemize}

\paragraph{Role Classifications (assign ONE per paper):}
\begin{enumerate}
\item \textbf{Baseline}: The primary system/method this paper improves upon or compares against as its main competitor
\item \textbf{Inspiration}: Paper whose specific idea/approach directly sparked the current paper's key innovation
\item \textbf{Gap Identification}: Paper whose explicit limitations/failures motivated this research direction
\item \textbf{Foundation}: Paper that introduced the core problem formulation, dataset, or theoretical framework used
\item \textbf{Extension}: Paper whose specific method is directly extended, modified, or generalized
\item \textbf{Related Problem}: Paper solving a closely related problem whose solution approach informed this work
\end{enumerate}

\paragraph{Output Requirements:}

For each prior work (identify 5-7 papers, quality over quantity):
\begin{enumerate}
\item \textbf{Role}: One of the six classifications above
\item \textbf{Relationship Sentence}: ONE specific sentence explaining the DIRECT connection to the current paper's innovation. Be concrete about WHAT was borrowed/extended/addressed.
\end{enumerate}

\paragraph{Synthesis Narrative (200-300 words):}

Write a cohesive narrative that flows naturally (NO explicit ``Part 1'' / ``Part 2'' labels):

\textbf{First $\sim$150 words - Prior Work with Relevant Details:}
Describe each prior work, but FOCUS ONLY on the specific aspects/details that relate to the current paper's innovation. For each prior work, highlight:
\begin{itemize}
\item The specific technique, insight, or finding that is relevant (not a general summary)
\item How this specific detail connects to what the current paper does
\item Do NOT mention the current paper yet---just establish what relevant knowledge existed
\end{itemize}

\textbf{Remaining $\sim$100 words - How They Collectively Inspired Current Work:}
Transition naturally to explain:
\begin{itemize}
\item What gap or opportunity emerged from the combination of these prior works
\item How the current paper synthesizes or builds upon these specific relevant details
\item Why this was a natural next step given the prior work landscape
\end{itemize}

The narrative should read as one flowing paragraph, not two separate sections.
\end{quote}

\textbf{User Prompt:}

\begin{quote}
Analyze this research paper and identify the prior works that DIRECTLY led to its core innovation.

\textbf{TASK:} Identify 5-7 prior works that DIRECTLY influenced this paper's KEY CONTRIBUTION.

Focus on papers that:
\begin{enumerate}
\item Introduced ideas/methods this paper directly builds on
\item Had limitations this paper explicitly addresses
\item Defined the problem formulation used here
\item Are the primary baselines being improved upon
\end{enumerate}

DO NOT include generic tools, orthogonal optimizations, or tangentially related work.

Return your analysis as valid JSON with the following
structure:

{\small
\begin{verbatim}
{
  "prior_works": [
    {
      "title": "Exact paper title",
      "authors": "First author et al.",
      "year": 2023,
      "arxiv_id": "if known",
      "role": "One of six roles",
      "relationship_sentence":
        "Specific sentence about DIRECT
         connection"
    }
  ],
  "synthesis_narrative":
    "200-300 word flowing narrative"
}
\end{verbatim}
}

Remember: Every paper you include should pass the test: ``This paper DIRECTLY influenced the core innovation, not just the general research area.''
\end{quote}

\subsection{Multi-Round Pattern Discovery Prompt}

\textbf{Purpose:} Discover patterns across multiple sampling rounds.

\textbf{System Prompt:}

\begin{quote}
You are an expert analyst of scientific innovation patterns. Always respond with valid JSON.
\end{quote}

\textbf{User Prompt Key Instructions:}

\begin{quote}
I have [N] synthesis narratives from top ML conference papers (ICML, ICLR, NeurIPS). Each narrative describes how authors built their novel contribution on prior work.

YOUR TASK: Identify the THINKING PATTERNS used by these researchers.

IMPORTANT:
\begin{itemize}
\item Look for COGNITIVE/REASONING patterns, not topic categories
\item Examples: ``Cross-domain analogy'', ``Constraint relaxation'', ``Theoretical unification'', ``Problem reframing'', ``Empirical observation leading to theory'', ``Modular decomposition''
\item Be specific and descriptive
\item Discover patterns inductively---don't limit to predefined categories
\end{itemize}
\end{quote}

\subsection{Pattern Consolidation Prompt}

\textbf{Purpose:} Consolidate discovered patterns into clean taxonomy.

\textbf{System Prompt:}

\begin{quote}
You are an expert in categorizing research innovation patterns. Create a rigorous taxonomy.
\end{quote}

\textbf{User Prompt Key Instructions:}

\begin{quote}
YOUR TASK: Consolidate these into a CLEAN TAXONOMY of 10-12 distinct, non-overlapping pattern categories.

Guidelines:
\begin{itemize}
\item Patterns that appear in more rounds are more robust/reliable
\item Merge semantically similar patterns
\item Each category should represent a distinct cognitive strategy
\item Avoid topic-based categories (focus on HOW researchers think, not WHAT they study)
\end{itemize}
\end{quote}

\subsection{LLM Judge Evaluation Prompt}
\label{app:llm-judge-prompt}

\textbf{Purpose:} Evaluate whether a generated research idea semantically matches a real published paper.

\textbf{System Prompt:}

\begin{quote}
You are evaluating whether a generated research idea matches a real published paper.

GENERATED IDEA:
Title: \{generated\_idea.title\}
Description: \{generated\_idea.description\}

REAL PUBLISHED PAPER:
Title: \{ground\_truth\_title\}
Contribution: \{ground\_truth\_contribution\}

Determine if the generated idea is semantically similar to the real paper. Consider:
\begin{enumerate}
\item Do they address the same core problem or research question?
\item Do they propose similar methodological approaches?
\item Would the generated idea, if fully developed, result in a similar contribution?
\end{enumerate}

A match means the ideas are substantially aligned in their core direction, not necessarily identical in every detail.

Respond with a JSON object containing:
\begin{itemize}
\item ``is\_match'': true or false
\item ``confidence'': a number from 0 to 1
\item ``reasoning'': a brief explanation (2-3 sentences)
\end{itemize}

Output ONLY the JSON object, no other text.
\end{quote}

\section{Annotation Schema and Guidelines}
\label{app:handbook}

This section provides excerpts from our annotation schema, which defines the structured categories and decision rules used by our LLM pipeline to classify predecessor roles and relationship types. These guidelines are incorporated into the LLM prompts to ensure consistent, high-quality structured annotations across all 3,819 papers. Note that these predecessor role and relationship type classifications are used for constructing the intellectual lineage graphs; they do not affect the subsequent thinking pattern analysis, which is derived independently from the synthesis narratives and innovation trajectories.

\subsection{Predecessor Role Definitions}

\textbf{KEY\_METHODOLOGY\_COMPONENT:} The source paper provides a specific technique, algorithm, model architecture, or methodological approach that is directly adopted or adapted by the target paper. This is concrete and technical.

\emph{Decision Rule:} If the target paper's implementation directly uses or builds upon a specific method from the source, annotate as KEY\_METHODOLOGY.

\emph{Example:} If the target paper uses RLHF as its core alignment method, then the original RLHF paper is a KEY\_METHODOLOGY\_COMPONENT.

\textbf{FOUNDATIONAL\_CONCEPT:} The source paper establishes theoretical grounding, introduces a fundamental concept, or provides the conceptual framework within which the target paper operates. This is more abstract than methodology.

\emph{Decision Rule:} If removing this source would make the target paper's theoretical motivation unclear or unfounded, annotate as FOUNDATIONAL\_CONCEPT.

\emph{Example:} If the target paper is about improving transformer efficiency, the original "Attention is All You Need" paper providing the transformer architecture is FOUNDATIONAL.

\textbf{PRIMARY\_BASELINE:} The source paper represents the state-of-the-art approach that the target paper aims to outperform or improve upon.

\emph{Decision Rule:} If the target paper explicitly compares its results against the source as the main point of comparison, annotate as PRIMARY\_BASELINE.

\textbf{PROBLEM\_FORMULATION:} The source paper defines, formalizes, or establishes the problem that the target paper addresses.

\emph{Decision Rule:} If the source introduced the task definition, dataset, or formal problem statement that the target paper tackles, annotate as PROBLEM\_FORMULATION.

\textbf{ENABLING\_TOOL\_OR\_DATASET:} The source provides a practical resource (codebase, dataset, benchmark) that the target paper uses.

\emph{Decision Rule:} If the source's primary contribution is a resource rather than an idea or method, and the target uses that resource, annotate as ENABLING\_TOOL.

\textbf{INSPIRATION\_BY\_ANALOGY:} The source paper comes from a different domain or problem area and provides inspiration through analogy or conceptual transfer.

\emph{Decision Rule:} If the source addresses a fundamentally different problem but the target adapts its approach, annotate as INSPIRATION\_BY\_ANALOGY.

\subsection{Relationship Type Definitions}

\textbf{EXTENDS:} The target directly builds upon, generalizes, or scales the source's approach in a relatively straightforward way.

\textbf{COMBINES\_WITH:} The target merges ideas from this source with ideas from other sources to create something new.

\emph{Decision Rule:} If the innovation comes from the synthesis of multiple distinct approaches, mark all relevant sources with COMBINES\_WITH.

\textbf{BRIDGES\_GAP\_BETWEEN:} The target connects this source with another, previously disconnected line of work.

\textbf{ADDRESSES\_LIMITATION\_OF:} The target explicitly identifies a shortcoming in the source and proposes a solution.

\textbf{REFRAMES\_USING:} The target reinterprets or reconceptualizes an existing problem using ideas from the source.

\section{Additional Case Studies}
\label{app:cases}

This section presents additional synthesis graphs illustrating how different thinking patterns manifest in recent high-impact work.

\subsection{Case Study: Flow Matching Generalization}

\noindent\textbf{Target Paper:} \emph{On the Closed-Form of Flow Matching: Generalization Does Not Arise from Target Stochasticity} (Bertrand et al., NeurIPS 2025 Oral)

\noindent\textbf{Core Contribution:} Demonstrates that stochasticity in conditional targets is not the primary driver of generalization in flow matching. Shows that closed-form deterministic velocity targets match or improve performance, and that generalization instead arises from the neural network's failure to perfectly approximate the optimal closed-form velocity field in specific time intervals.

\paragraph{Intellectual Predecessors}

This work synthesizes five distinct research lineages:

\begin{enumerate}
\item \textbf{Flow Matching for Generative Modeling} (Lipman et al., 2023) --- \textsc{Key Methodology Component}. Introduced the conditional flow matching objective and practical algorithms for learning time-dependent velocity fields that transport a simple prior into the data distribution. Provided the training loss, sampling ODE, and experimental setup that the target paper studies and modifies.

\item \textbf{Score-Based Generative Modeling} (Song et al., 2021) --- \textsc{Foundational Concept}. Established the SDE/score-matching perspective linking noisy conditional training objectives to continuous-time generative dynamics. This conceptual lens frames questions about the role of stochasticity in training targets and motivates comparing noisy sampled targets to deterministic closed-form targets.

\item \textbf{Foundational Flow/Transport Theory} (Albergo \& Vanden-Eijnden, 2023) --- \textsc{Key Methodology Component}. Developed theoretical tools and variants of conditional flow/bridge formulations, including derivations yielding closed-form velocity fields. These results enable construction and computation of deterministic closed-form targets that replace sampled stochastic targets.

\item \textbf{Empirical Study of Memorization vs Generalization in Diffusion Models} (Kadkhodaie et al., 2024) --- \textsc{Problem Formulation}. Provided empirical evidence that diffusion models can either memorize or generalize depending on dataset size, model capacity, and regime. This empirical context motivated the central question: what mechanisms cause flow matching models to generalize in realistic regimes?

\item \textbf{Noisy Training-Loss Explanation for Generalization} (Vastola, 2025) --- \textsc{Inspiration by Analogy}. Hypothesized that the stochastic nature of conditional training targets induces implicit regularization promoting generalization. The target paper tests and challenges this hypothesis specifically in the flow matching setting.
\end{enumerate}

\paragraph{Key Inter-Predecessor Relationships}

\begin{itemize}
\item Song et al.'s SDE/score viewpoint \textbf{bridges the gap between} diffusion and Lipman et al.'s flow matching, enabling comparisons of stochastic vs deterministic training targets.
\item Albergo \& Vanden-Eijnden's closed-form/bridge theory \textbf{combines with} Lipman et al.'s flow-matching framework, producing concrete closed-form alternatives to sampled conditional targets.
\item Kadkhodaie et al.'s empirical findings \textbf{address limitations of} Vastola's hypothesis by exposing the need for mechanistic explanations of regime-dependent memorization/generalization.
\item Vastola \textbf{reframes} successes of Lipman et al.'s methods through the lens of noisy training losses, prompting a targeted test within the flow matching setup.
\end{itemize}

\paragraph{Thinking Trajectory}

\emph{Starting point:} Researchers worked within the flow-matching paradigm using conditional flow matching loss to learn time-dependent velocity fields.

\emph{Gap identified:} Existing explanations for why these models generalize were incomplete. Empirical studies showed regime-dependent memorization, and a prominent hypothesis claimed that noisy/stochastic training targets provided implicit regularization---but this had not been tested in high-dimensional, realistic flow-matching settings.

\emph{Key insight:} Closed-form optimal conditional velocity fields can be computed from transport/bridge theory, enabling direct testing of whether target stochasticity is essential for generalization by replacing sampled stochastic targets with deterministic closed-form targets.

\emph{Reframing:} Rather than treating stochasticity as an inescapable feature of training, the authors reframed the question as a controlled comparison between stochastic (sampled) targets and their closed-form deterministic counterparts within the same flow-matching framework.

\emph{Novel combination:} Combined Lipman et al.'s practical flow-matching training and architectures with Albergo \& Vanden-Eijnden's closed-form velocity expressions, and designed empirical protocols inspired by Kadkhodaie et al.'s regime analyses to evaluate Vastola's noisy-loss hypothesis under realistic, high-dimensional conditions.

\emph{Resulting innovation:} A conclusive empirical and diagnostic demonstration that stochastic target noise is not the key driver of generalization in flow matching. Instead, generalization arises when limited-capacity networks fail to approximate the optimal closed-form velocity field in specific time intervals, and closed-form targets can match or improve performance.

\paragraph{Patterns Exemplified}

This work demonstrates several thinking patterns:
\begin{itemize}
\item \textbf{Gap-Driven Reframing:} Identified the gap between empirical generalization behavior and mechanistic explanations, then reframed the question as a testable hypothesis about target stochasticity.
\item \textbf{Cross-Domain Synthesis:} Combined insights from transport theory, score-based diffusion models, and empirical regime analyses to create a unified experimental framework.
\item \textbf{Formal-Experimental Tightening:} Paired theoretical closed-form derivations with rigorous empirical validation across multiple datasets and model scales.
\end{itemize}

\begin{table}[t]
  \centering
  \small
  \begin{tabular}{lr}
  \toprule
  \textbf{Thinking Pattern} & \textbf{\%} \\
  \midrule
  Gap-Driven Reframing & 24.2 \\
  Cross-Domain Synthesis & 18.0 \\
  Representation Shift \& Primitive Recasting & 10.5 \\
  Formal-Experimental Tightening & 6.7 \\
  Principled Probabilistic Modeling & 5.4 \\
  Data \& Evaluation Engineering & 5.4 \\
  Inject Structural Inductive Bias & 5.1 \\
  Approximation Engineering for Scalability & 4.9 \\
  Modular Pipeline Composition & 4.2 \\
  Inference-Time Control \& Guided Sampling & 2.4 \\
  Data-Centric Optimization \& Active Sampling & 2.1 \\
  Adversary Modeling \& Defensive Repurposing & 1.5 \\
  Multiscale \& Hierarchical Modeling & 1.4 \\
  Self-Supervised Pretext Engineering & 1.9 \\
  Meta-Learning \& Learning-to-Learn & 1.5 \\
  \bottomrule
  \end{tabular}
  \caption{Complete distribution of thinking patterns across all 3,819 analyzed papers.}
  \label{tab:full-patterns}
  \end{table}

\section{Extended Statistical Analysis}
\label{app:stats}

This section provides additional statistical details from our analysis of 3,819 papers from NeurIPS, ICML, and ICLR (2023--2025).

\subsection{Dataset Collection Notes}
\label{app:dataset-notes}

The paper counts in Figure~\ref{fig:dataset-distribution} may differ from official conference statistics for two reasons. First, some papers were unavailable at the time of collection due to access restrictions, or missing files. Second, we focus specifically on research contributions and exclude certain paper types from our analysis, including benchmark papers, dataset papers, technical reports, position papers, and survey papers. These exclusions ensure that our analysis of thinking patterns reflects methodological and conceptual innovations rather than resource contributions or meta-analyses. The final dataset of 3,819 papers represents the intersection of papers that were both accessible and classified as research contributions.

\subsection{Complete Pattern Frequency Distribution}

Table~\ref{tab:full-patterns} presents the complete distribution of all 15 thinking patterns identified in our taxonomy, ordered by frequency.

\subsection{Pattern Co-occurrence Analysis}

\begin{figure}[t]
\centering
\includegraphics[width=\columnwidth]{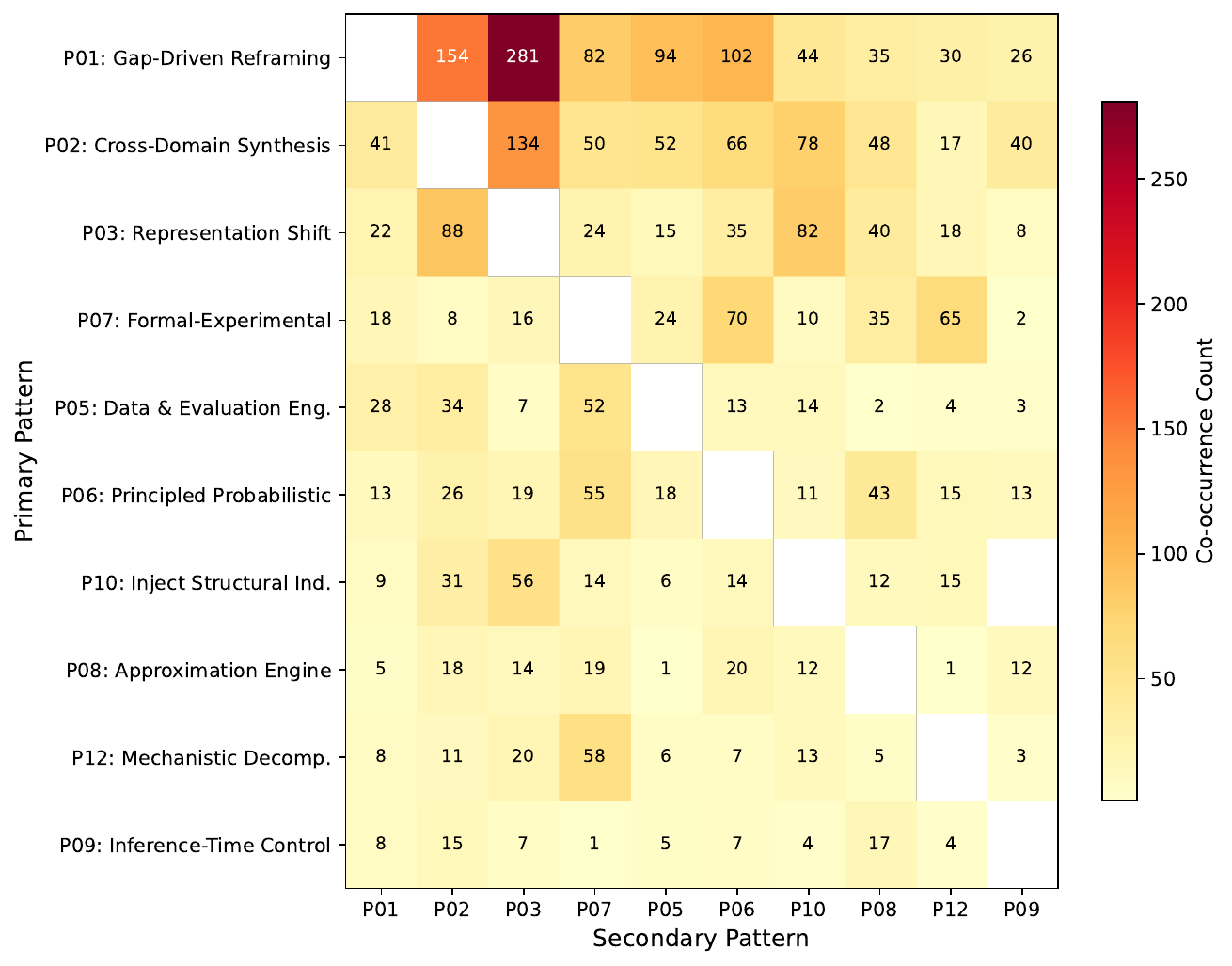}
\caption{Pattern co-occurrence heatmap. Darker cells indicate more frequent co-occurrence. The diagonal represents single-pattern papers.}
\label{fig:cooccurrence}
\end{figure}

Figure~\ref{fig:cooccurrence} shows the co-occurrence heatmap revealing which thinking patterns frequently appear together. The strongest co-occurrences include:

\begin{itemize}
    \item \textbf{Gap-Driven Reframing + Representation Shift} (318 co-occurrences): This ``reframe+repr'' combination represents the canonical path to breakthrough work---identifying a limitation and resolving it through primitive changes.
    \item \textbf{Cross-Domain Synthesis + Representation Shift} (233): Importing methods from other domains often requires adapting representations to the target setting.
    \item \textbf{Gap-Driven Reframing + Cross-Domain Synthesis} (204): Gaps are frequently addressed by borrowing solutions from adjacent fields.
    \item \textbf{Representation Shift + Inject Structural Inductive Bias} (145): New representations often incorporate domain-specific structure.
    \item \textbf{Principled Probabilistic Modeling + Formal-Experimental Tightening} (131): Theoretical work pairs naturally with rigorous validation.
\end{itemize}

\subsection{Oral vs. Spotlight Presentation Analysis}

\begin{figure}[ht]
\centering
\includegraphics[width=\columnwidth]{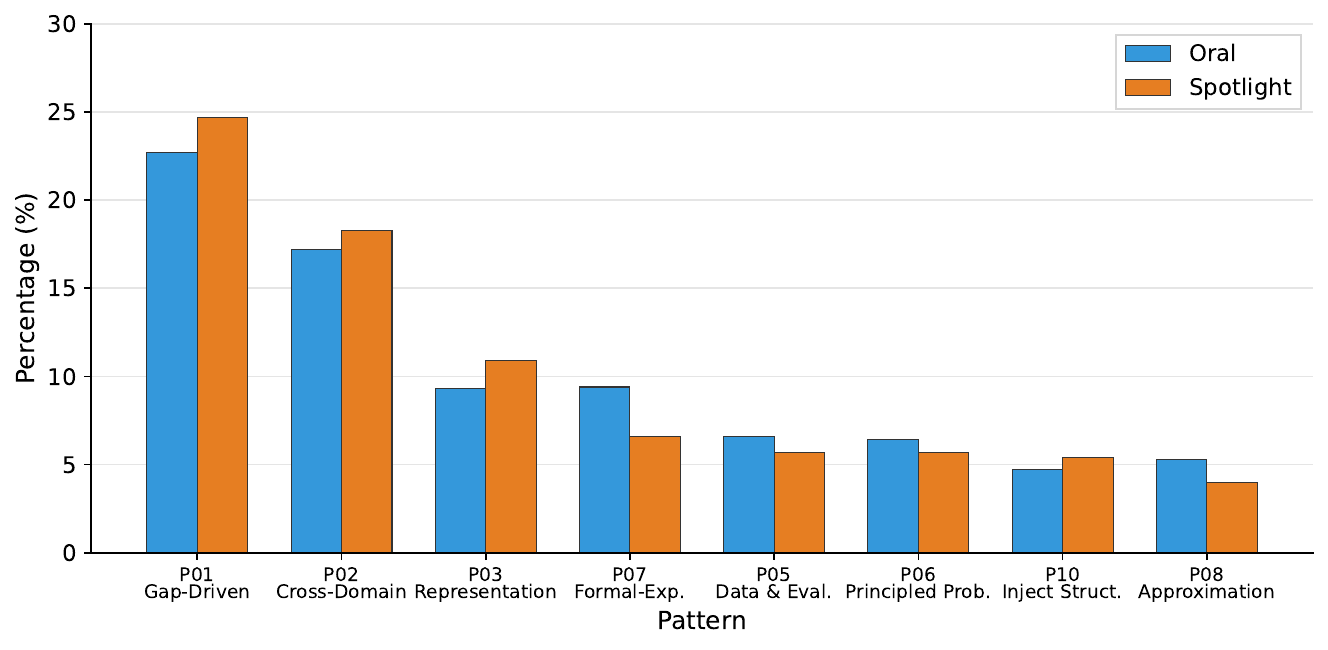}
\caption{Distribution of thinking patterns across oral and spotlight presentations.}
\label{fig:oral-spotlight}
\end{figure}

Our dataset contains 999 oral presentations and 2,820 spotlight presentations. Figure~\ref{fig:oral-spotlight} shows the pattern distribution by presentation type.

Key observations include:
\begin{itemize}
    \item Papers receiving oral presentations show higher concentrations of Gap-Driven Reframing combined with Representation Shift---the ``reframe+repr'' recipe correlates with maximum visibility.
    \item Cross-Domain Synthesis with Modular Pipeline Composition (106 co-occurrences overall) appears frequently in oral presentations, suggesting that work demonstrating broad applicability receives higher recognition.
    \item The combination of Principled Probabilistic Modeling with Formal-Experimental Tightening predicts durable technical influence, appearing in papers that receive sustained citations.
\end{itemize}

\subsection{Detailed Temporal Evolution}

Beyond the year-over-year trends presented in the main text, we observe the following detailed temporal patterns:

\textbf{Stable Patterns:} Gap-Driven Reframing remains remarkably consistent (26.1\% $\rightarrow$ 23.7\% $\rightarrow$ 23.8\% from 2023--2025), indicating a persistent research mode of ``diagnose $\rightarrow$ reframe.''

\textbf{Rising Patterns:} Representation Shift showed a notable increase in 2024 (8.0\% $\rightarrow$ 11.5\%), reflecting a wave of representational innovations including new primitives, modalities, and implicit representations. Data \& Evaluation Engineering shows modest growth in 2025 (6.6\%), suggesting increased attention to benchmarks and reproducibility.

\textbf{Declining Patterns:} Formal-Experimental Tightening decreased as a primary pattern (10.1\% $\rightarrow$ 7.1\% $\rightarrow$ 6.6\%), suggesting formalization is increasingly a supporting rather than headline contribution.

\subsection{Conference-Specific Detailed Statistics}

\textbf{ICLR:} Shows stronger emphasis on Representation Shift (11.8\%) and Data \& Evaluation Engineering (8.5\%), consistent with ICLR's reputation for representation learning and empirical benchmarks.

\textbf{ICML:} Higher concentrations of Gap-Driven Reframing (25.8\%), Formal-Experimental Tightening (8.3\%), and Principled Probabilistic Modeling (7.5\%), aligning with ICML's tradition of statistically grounded algorithmic work.

\textbf{NeurIPS:} Broadly balanced distribution with Gap-Driven Reframing (24.5\%), Cross-Domain Synthesis (18.5\%), and Formal Tightening (8.1\%), reflecting NeurIPS's cross-disciplinary character.

\subsection{Underexplored Opportunity Spaces}

Patterns with low primary frequency represent fertile areas for high-impact contributions:

\begin{itemize}
    \item \textbf{Multiscale \& Hierarchical Modeling} (1.5\%): Many real systems exhibit hierarchical structure; deeper development could yield efficiency and interpretability gains.
    \item \textbf{Data-Centric Optimization \& Active Sampling} (2.3\%): Growing interest in data efficiency makes explicit active sampling methods increasingly relevant.
    \item \textbf{Inference-Time Control \& Guided Sampling} (2.7\%): Systems that adapt at inference to trade off compute and quality are underexplored relative to deployment value.
    \item \textbf{Adversary Modeling \& Defensive Repurposing} (1.7\%): Security and robustness patterns remain small but will become strategically important as ML deployments scale.
\end{itemize}

\subsection{Actionable Insights for Researchers}

The identified patterns translate directly into actionable strategies for conducting research. For PhD students and early-career researchers, the data suggests starting with focused ``gap identification'' exercises: systematically analyzing recent papers to write explicit gap statements reveals opportunities that others may have overlooked. Mastering at least one tool or formalism from an adjacent field (control theory, probabilistic graphical models, implicit representations, optimization theory) provides the raw material for cross-domain synthesis. The most accessible entry point is the ``Reframe + Represent'' trajectory: identify a crisp limitation, ask ``what primitive would make this simple?'', and prototype with the new representation.

For experienced researchers and research teams, the analysis suggests investing in ``reframe + representation'' projects backed by rigorous validation, building cross-domain collaborations that pair empiricists with theoreticians and domain experts, and creating transferable tooling around innovations to maximize impact. The temporal trends also reveal underexplored opportunities: patterns like Inference-Time Control (2.7\%), Multiscale Hierarchical Modeling (1.5\%), and Adversarial Robustness (1.7\%) show low current adoption but high potential impact, particularly for industry deployment scenarios. Finally, the conference-specific patterns suggest tailoring submission strategies: ICML submissions should emphasize mathematical rigor and statistical foundations, ICLR submissions should highlight architectural innovations and benchmark contributions, while NeurIPS submissions benefit from broad applicability and cross-disciplinary synthesis.

\section{Complete List of 15 Thinking Patterns}
\label{app:complete-patterns}

This section provides detailed descriptions of all 15 thinking patterns identified in our analysis of 3,819 papers. Each pattern includes its cognitive strategy, key indicators, concrete examples, and actionable insights for researchers.

\subsection{P01: Gap-Driven Reframing (24.2\%)}

\noindent\textbf{Cognitive Move:} Turn a specific failure or mismatched assumption into an explicit design constraint that maps the problem onto better-suited methods.

\noindent\textbf{Example:} Reframing autoregressive image modeling from next-token prediction to next-scale (coarse→fine) prediction.

\noindent\textbf{Learnable Insight:} When you notice a recurring failure, write it as an explicit constraint; ask ``if this limitation were the problem, what methods would apply?''

\subsection{P02: Cross-Domain Synthesis (18.0\%)}

\noindent\textbf{Cognitive Move:} Map components across disciplinary boundaries and transplant them while engineering the compatibility layer.

\noindent\textbf{Example:} Fusing quantum circuits with transformer attention to obtain doubly stochastic attention matrices.

\noindent\textbf{Learnable Insight:} List constraints your method fails to satisfy, search other fields for primitives addressing those constraints, and prototype with a thin adapter.

\subsection{P03: Representation Shift \& Primitive Recasting (10.5\%)}

\noindent\textbf{Cognitive Move:} Replace the problem's language (pixels, tokens, meshes) with an alternative primitive that simplifies inference or constraints.

\noindent\textbf{Example:} Replacing explicit meshes with neural implicit signed-distance functions for 3D reconstruction.

\noindent\textbf{Learnable Insight:} When a task struggles with geometry or combinatorics, enumerate alternative primitives and test whether the new one reduces complexity.

\subsection{P04: Modular Pipeline Composition (4.7\%)}

\noindent\textbf{Cognitive Move:} Decompose a complex end-to-end system into composable modules with clean interfaces, enabling mix-and-match flexibility.

\noindent\textbf{Example:} Breaking monolithic vision models into separate perception, reasoning, and generation modules that can be independently upgraded.

\noindent\textbf{Learnable Insight:} When facing a complex system, identify natural factorization points and design interfaces that allow independent optimization of components.

\subsection{P05: Data \& Evaluation Engineering (6.0\%)}

\noindent\textbf{Cognitive Move:} Create new datasets, benchmarks, or evaluation metrics that expose previously unmeasured phenomena or enable fairer comparisons.

\noindent\textbf{Example:} Designing contamination-resistant evaluation protocols that detect when models have memorized test data during pretraining.

\noindent\textbf{Learnable Insight:} When existing benchmarks saturate or fail to capture important behaviors, invest in principled evaluation infrastructure as a research contribution.

\subsection{P06: Principled Probabilistic Modeling (6.0\%)}

\noindent\textbf{Cognitive Move:} Formulate the problem using probabilistic graphical models, Bayesian inference, or statistical principles that provide interpretability and theoretical guarantees.

\noindent\textbf{Example:} Replacing heuristic neural architectures with structured variational autoencoders that factorize latent representations according to causal structure.

\noindent\textbf{Learnable Insight:} When a problem involves uncertainty, composition, or interpretability requirements, consider whether probabilistic frameworks provide cleaner solutions than deterministic neural networks.

\subsection{P07: Formal-Experimental Tightening (7.4\%)}

\noindent\textbf{Cognitive Move:} Iterate between theoretical analysis (proofs, convergence guarantees, sample complexity) and empirical validation, using each to refine the other.

\noindent\textbf{Example:} Proving sample complexity bounds for a reinforcement learning algorithm, then using ablations to verify that empirical performance matches theoretical predictions.

\noindent\textbf{Learnable Insight:} Strong papers pair conceptual innovations with rigorous validation---either formal guarantees or exhaustive empirical analysis that rules out confounds.

\subsection{P08: Approximation Engineering for Scalability (5.4\%)}

\noindent\textbf{Cognitive Move:} Replace exact but intractable operations with principled approximations (sketching, quantization, low-rank, sparse) that preserve essential properties while enabling scale.

\noindent\textbf{Example:} Approximating full attention with locality-sensitive hashing to achieve subquadratic complexity while maintaining quality.

\noindent\textbf{Learnable Insight:} When hitting computational bottlenecks, identify which properties are essential for downstream performance and design approximations that preserve those while discarding expensive-but-inessential structure.

\subsection{P09: Inference-Time Control \& Guided Sampling (2.7\%)}

\noindent\textbf{Cognitive Move:} Enable runtime control over model behavior through guided sampling, inference-time optimization, or adaptive computation without retraining.

\noindent\textbf{Example:} Using classifier-free guidance to steer diffusion models toward desired attributes at generation time without fine-tuning.

\noindent\textbf{Learnable Insight:} When deployment requirements vary (quality-latency tradeoffs, diverse user preferences), design systems that adapt at inference time rather than requiring separate fine-tuned models.

\subsection{P10: Inject Structural Inductive Bias (5.7\%)}

\noindent\textbf{Cognitive Move:} Hardcode domain-specific structure (symmetries, invariances, geometric priors) into architectures to improve sample efficiency and generalization.

\noindent\textbf{Example:} Designing graph neural networks with permutation equivariance to handle variable-sized molecular structures.

\noindent\textbf{Learnable Insight:} Identify structural properties that hold universally in your domain (rotation, permutation, scaling invariances) and bake them into architectures as hard constraints.

\subsection{P11: Multiscale \& Hierarchical Modeling (1.5\%)}

\noindent\textbf{Cognitive Move:} Explicitly model phenomena at multiple levels of granularity or abstraction, with cross-scale interactions.

\noindent\textbf{Example:} Processing images through a pyramid of resolutions, with top-down and bottom-up information flow between levels.

\noindent\textbf{Learnable Insight:} When systems exhibit natural hierarchies (visual scenes, language semantics, physical simulations), explicit multiscale modeling often outperforms single-scale approaches.

\subsection{P12: Mechanistic Decomposition \& Causal Localization (3.8\%)}

\noindent\textbf{Cognitive Move:} Decompose model behavior into interpretable mechanisms or localize causal effects to specific components through interventions.

\noindent\textbf{Example:} Using activation patching to identify which transformer attention heads are causally responsible for specific reasoning capabilities.

\noindent\textbf{Learnable Insight:} When models exhibit surprising behaviors, systematic decomposition and causal interventions reveal which components are necessary and sufficient.

\subsection{P13: Adversary Modeling \& Defensive Repurposing (1.7\%)}

\noindent\textbf{Cognitive Move:} Model potential adversaries or failure modes explicitly, then design defenses or repurpose adversarial techniques for robustness.

\noindent\textbf{Example:} Using adversarial training to improve model robustness, or detecting out-of-distribution inputs through learned adversarial perturbations.

\noindent\textbf{Learnable Insight:} Explicitly modeling worst-case scenarios and adversarial behaviors leads to more robust systems than hoping for benign deployment conditions.

\subsection{P14: Numerics \& Systems Co-design (1.4\%)}

\noindent\textbf{Cognitive Move:} Co-design algorithms and their low-level implementation (numerical precision, memory layout, kernel fusion) to achieve orders-of-magnitude speedups.

\noindent\textbf{Example:} Redesigning transformer operations to exploit tensor core hardware, achieving 10x speedups through algorithmic and systems co-design.

\noindent\textbf{Learnable Insight:} For deployment-critical systems, collaborate with systems experts to co-design algorithms that exploit hardware characteristics rather than treating implementation as an afterthought.

\subsection{P15: Data-Centric Optimization \& Active Sampling (2.3\%)}

\noindent\textbf{Cognitive Move:} Optimize the data distribution rather than (or in addition to) the model, using active learning, curriculum design, or strategic data curation.

\noindent\textbf{Example:} Using uncertainty-based active learning to identify maximally informative unlabeled examples, reducing labeling requirements by 10x.

\noindent\textbf{Learnable Insight:} When data acquisition is expensive or noisy, strategic data selection and curation can yield larger gains than architectural improvements.

\end{document}